\pdfoutput=1

\documentclass[11pt]{article}

\usepackage{acl}

\usepackage{times}
\usepackage{latexsym}

\usepackage[T1]{fontenc}

\usepackage[utf8]{inputenc}

\usepackage{microtype}

\usepackage{inconsolata}

\usepackage{booktabs}        
\usepackage{amsmath}         
\usepackage{amssymb}         
\usepackage{amsfonts}        
\usepackage{autobreak}       
\usepackage{mathtools}       
\usepackage{multirow}        
\usepackage{bbm}             
\usepackage{graphicx,xcolor} 
\usepackage{pxrubrica}       
\usepackage{color}           
\usepackage{xcolor}          
\usepackage{algorithm}       
\usepackage{algpseudocode}   
\usepackage[normalem]{ulem}  
\usepackage{url}
\usepackage{comment}

\def\vec#1{\ensuremath{\boldsymbol{#1}}} 
\def\v#1{\ensuremath{\vec{v}_{\mathit{#1}}}} 
\newcommand{\w}[1]{\ensuremath{\mathit{#1}}}  
\newcommand{\set}[1]{\ensuremath{\mathit{#1}}}  
\newcommand{\Span}[1]{\ensuremath{\mathbb{#1}}}  
\newcommand{\Subs}[1]{\ensuremath{\Span{S}_{\mathit{#1}}}}  
\newcommand{\m}[1]{\boldsymbol{\mathrm{#1}}}  
\newcommand{\ms}[1]{\ensuremath{\m{S}_{\mathrm{#1}}}}  

\usepackage{pifont}
\newcommand{\pifontcheckmark}{\text{\ding{52}}}
\newcommand{\pifontcrossmark}{\text{\ding{56}}}
\newcommand{\chk}{\textcolor{teal}{\pifontcheckmark}}
\newcommand{\x}{\textcolor{magenta}{\pifontcrossmark}}

\newcommand{\hardin}{\mathbbm{1}_{\mathrm{hard}}} 
\newcommand{\softin}{\mathbbm{1}_{\mathrm{subspace}}} 
 
\newcommand{\mem}[1]{\mathbbm{1}_{\mathrm{#1}}}

\newcommand{\blue}[1]{\colorbox[rgb]{0.92,0.96,1}{#1}} 
 
\newcommand{\gray}[1]{\colorbox[rgb]{0.9,0.9,0.9}{#1}}

\DeclarePairedDelimiter\abs{\lvert}{\rvert}

\usepackage{amsmath}
\usepackage[english]{babel}
\usepackage{hyperref}
\addto\captionsenglish{%
}
\addto\extrasenglish{%
}

\title{Subspace Representations for Soft Set Operations and Sentence Similarities}

\author{
   Yoichi Ishibashi$^{1,2}$\thanks{~~Work done while at Nara Institute of Science and Technology.} \hspace{0.7cm} Sho Yokoi$^{3, 4}$ \hspace{0.7cm} Katsuhito Sudoh$^{1}$ \hspace{0.7cm} Satoshi Nakamura$^{1}$
   \\
   $^1$ Nara Institute of Science and Technology \hspace{0.7cm} $^2$ Kyoto University \\\hspace{0.7cm} $^3$ Tohoku University \hspace{0.7cm} $^4$ RIKEN
   \\
   \texttt{\{ishibashi.yoichi.ir3, sudoh, s-nakamura\}@is.naist.jp}
   \\
   \texttt{yokoi@tohoku.ac.jp}
}

\begin{document}
\maketitle
\begin{abstract}
In the field of natural language processing (NLP), continuous vector representations are crucial for capturing the semantic meanings of individual words.
Yet, when it comes to the representations of \emph{sets} of words, the conventional vector-based approaches often struggle with expressiveness and lack the essential set operations such as union, intersection, and complement.
Inspired by quantum logic, we realize the representation of word sets and corresponding set operations within pre-trained word embedding spaces.
By grounding our approach in the linear subspaces, we enable efficient computation of various set operations and facilitate the \emph{soft} computation of membership functions within continuous spaces. Moreover, we allow for the computation of the F-score directly within word vectors, thereby establishing a direct link to the assessment of sentence similarity.
In experiments with widely-used pre-trained embeddings and benchmarks, we show that our subspace-based set operations consistently outperform vector-based ones in both sentence similarity and set retrieval tasks.
\footnote{Our code is publicly available at \url{https://github.com/yoichi1484/subspace}}
\end{abstract}

\section{Introduction}
Embedding-based word representations have become fundamental in the field of natural language processing (NLP). 
Models like word2vec~\cite{DBLP:conf/nips/MikolovSCCD13} and GloVe~\cite{DBLP:conf/emnlp/PenningtonSM14}, along with recent Transformer-based architectures~\cite{DBLP:conf/nips/VaswaniSPUJGKP17, DBLP:conf/naacl/DevlinCLT19}, have underscored the significance of embeddings in capturing the complexities of linguistic semantics.

The importance of representing collections of words is pivotal in understanding concepts and relationships within language contexts~\cite{DBLP:conf/nips/ZaheerKRPSS17,DBLP:conf/iclr/ZhelezniakSSMFH19}. 
For instance, while words like ``apple'' and ``orange'' each carry their distinct meanings, together they represent the broader concept of fruits. Another example of important application is a sentence representation~\cite{DBLP:conf/nips/ZaheerKRPSS17}. 
The set of words in a sentence captures the overall meaning, allowing for computations such as text similarity~\cite{DBLP:conf/semeval/AgirreCDG12}.

Against this backdrop, our research recognizes the significance of applying set operations in NLP and explores a new approach. 
Set operations enable a richer representation of relationships between collections of words, leading to more accurate semantic analysis based on context. 
For example, employing set operations allows for a clearer understanding of shared semantic features and differences among word groups within a text. 
This directly benefits tasks like determining semantic similarity and expanding word sets.

\begin{figure}[t]
\centering
\includegraphics[clip, width=8cm]{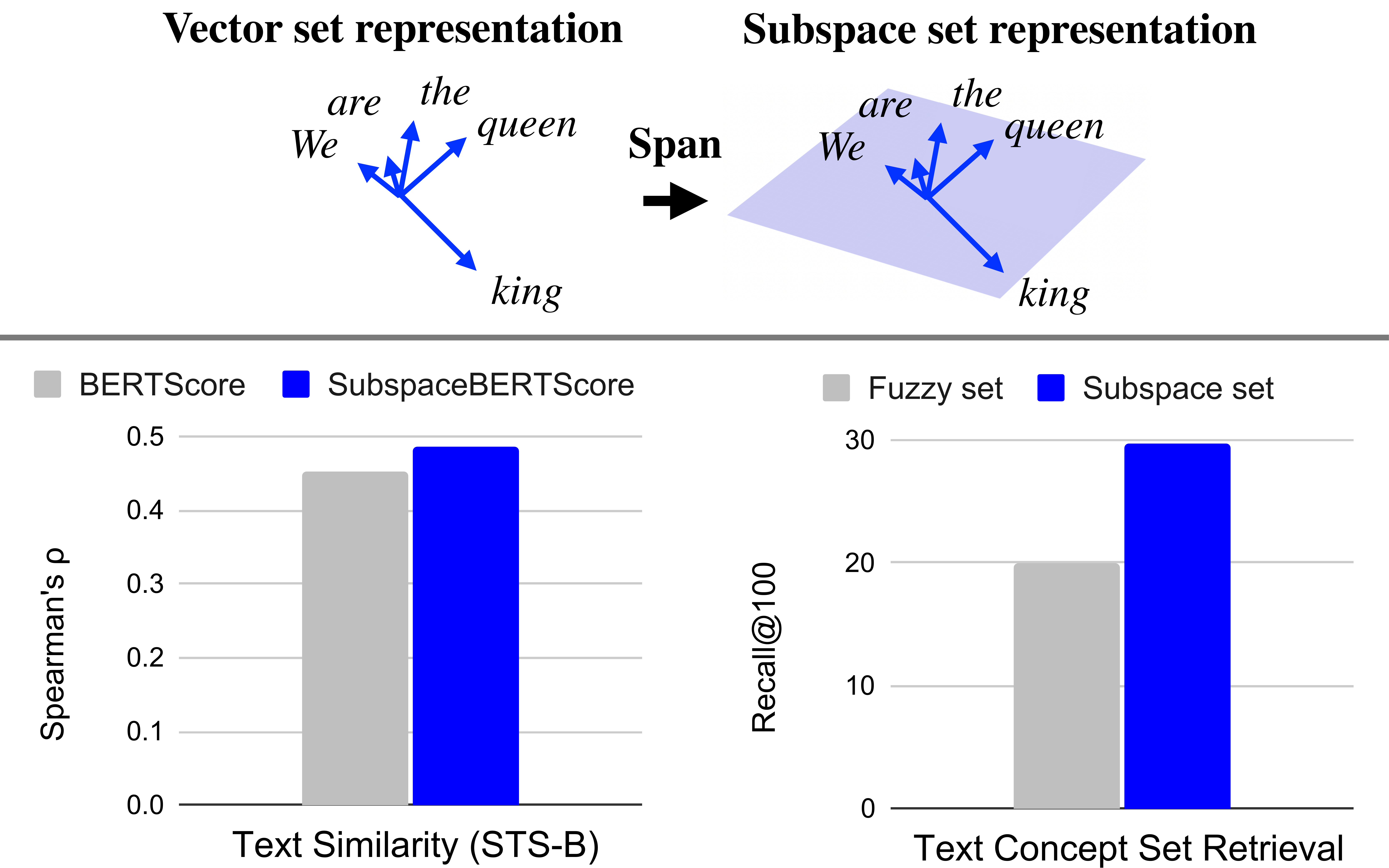}
\caption{Superiority of subspace representations: Our \blue{subspace representation (blue)} surpasses the traditional \gray{vector set representation (gray)} in both text similarity and text concept set retrieval tasks.}
\label{fig:intro_graph}
\end{figure}

In response to these challenges, our study introduces a novel methodology that exploits the principles of \emph{quantum logic}~\cite{birkhoff1936logic}, applied within embedding spaces to define set operations. 
Our proposed framework adopts a subspace-based approach for representing word sets, aiming to maintain the intricate semantic relationships within these sets. 
We represent a word set as a subspace which is spanned by pre-trained embeddings.
Additionally, it adheres to the foundational laws of set theory as delineated in the framework of quantum logic. 
This compliance ensures that our set operations, such as union, intersection, and complement, are not only mathematically robust but also linguistically meaningful when applied them in pre-trained embedding space. 

We first introduce a subspace set representation along with basic operations ($\cap$, $\cup$, and $\in$).
Subsequently, to highlight the usefulness of our proposed framework, we introduce two core set computations: text similarity and set membership.
The empirical results consistently point towards the notable superiority of our approach; our straightforward approach of spanning subspaces with pre-trained embedding sets enables a rich set representation, and we demonstrated its consistent performance enhancement in downstream tasks (\autoref{fig:intro_graph}).
Our research contributions include:
\begin{enumerate}
    \item 
    The introduction of continuous set representations and a framework for set operations, enabling more effective manipulation of word embedding sets (\autoref{sec:quantum}).

    \item 
    We propose \emph{SubspaceBERTScore}, an extension of the embedding set-based text similarity method, BERTScore~\cite{DBLP:conf/iclr/ZhangKWWA20}. 
    By simply transitioning from a vector set representation to a subspace, and incorporating a subspace-based indicator function, we observe a salient improvement in performance across all text similarity benchmarks (\autoref{sec:sts}). 
    
    \item
    We apply subspace-based basic operations ($\cap, \cup,$ and $\in$) to set expansion task and achive high performance (\autoref{sec:tcsr}). 
\end{enumerate}

\section{Preliminaries}
\label{sec:pre}
To make the following discussion clear, we define several symbols.
The sets of tokens in two sentences ($A$ and $B$) are denoted as $A = \{a_1, a_2, \dots\}, B = \{b_1, b_2, \dots\}$ respectively.
The sets of contextualized token vectors are denoted as $\m{A} = \{\vec{a}_1, \vec{a}_2, \dots\}, \m{B} = \{\vec{b}_1, \vec{b}_2, \dots\}$, where $\vec{a}$ and $\vec{b}$ are token vectors generated by the pre-trained embedding model such as BERT.
The subspace spanned by $\m{A}$ is denoted as $\Subs{A} = \mathrm{span}(\vec{a}_1, \vec{a}_2, \dots)$.
Note that the bases of the subspace is orthonormalized.

\section{Symbolic Set Operations} \label{sec:problem}
We first formulate various set operations in a pre-trained embedding space.
Among many types of operations for practical NLP applications, this work focuses on \emph{set similarity}:
\begin{equation}
\begin{split}
    & \set{A} = \{\w{A}, \w{boy}, \w{walks}, \w{in}, \w{this}, \w{park}\}\text{,}
    \\
    & \set{B} = \{\w{The}, \w{kid}, \w{runs}, \w{in}, \w{the}, \w{square}\}\text{,}
    \\
    & \mathrm{Similarity}(\set{A}, \set{B})\text{,}
    \label{eq:set_sim}
\end{split}
\end{equation}
\emph{set membership} ($\in$) and \emph{basic operations} ($\cap$, $\cup$):
\begin{equation}
\begin{split}
    & \set{Color} = \{\w{red}, \w{blue}, \w{green}, \w{orange}, \dots\}\text{,}
    \\
    & \set{Fruit} = \{\w{apple}, \w{orange}, \w{peach}, \dots\}\text{,}
    \\ 
    & \w{orange} \in \set{Color} \cap \set{Fruit}\text{.}
    \label{eq:conceptset}
\end{split}
\end{equation}

For this purpose, we need following representations on a pre-trained embedding space\footnote{These operations do not include some set operations such as cardinality, but are sufficient for expressing the practical forms of sets such as Eq. \eqref{eq:set_sim}.}:

\paragraph{An element and a set of elements}
The representations of an element and a set of elements are the most basic ones.
To exploit word embeddings, we represent a word (e.g., $\w{orange}$) as an element and a group of words (e.g., $\{\w{red}, \w{blue}, \w{green}, \w{orange}, \dots \}$) as a word set.

\paragraph{Quantification of set membership (indicator function)}
\emph{Membership} denotes a relation in which word $\w{w}$ is an element of set $\set{A}$, i.e., $\w{w}$ \blue{$\in$} $\set{A}$.
We quantify it based on vector representations.
Although the membership is typically a binary decision identical to that in a symbolic space, it can also be measured by the degree of closeness in a continuous vector space.
Membership can be computed as an indicator function.
The indicator function $\mem{set}$ quantifies whether the word $\w{w}$ is included ($1$) or not ($0$) in the set in a discrete manner:
\begin{align}
    \mem{set}[\w{w} \in \set{A}] =
    \begin{cases}
        1 & \mathrm{if} ~~\w{w} \in \set{A}\text{,} \\
        0 & \mathrm{if} ~~\w{w} \notin \set{A}\text{.}
    \end{cases}
    \label{eq:set-indictor}
\end{align}

\paragraph{Similarity between discrete symbol sets}
Set similarity, such as recall and precision, is an essential operation when calculating the similarity of texts.
Despite its simplicity, the word overlap-based sentence similarity serves as a remarkably effective approximation and has found widespread practical application, as evidenced by numerous studies\cite{DBLP:conf/wmt/BojarFFGHKM18,DBLP:conf/iclr/ZhangKWWA20,DBLP:conf/semeval/CerDALS17,DBLP:conf/iclr/ZhelezniakSSMFH19}.
They stand out as excellent similarity metrics based on embeddings. 
BERTScore~\cite{DBLP:conf/iclr/ZhangKWWA20}, which utilizes embeddings for its computation, is grounded in recall and precision~\footnote{Unlike the symbolic set similarity, which do not consider word order, contextualized embeddings enable the capture of word sequence information.}. 
The typical computations for recall ($R$) and precision ($P$) are as follows\footnote{For simplicity in explanation, we present $A$ and $B$ in Eq. \eqref{eq:exact-R} and Eq. \eqref{eq:exact-P} as a set of tokens.}:
\begin{align}
    R &= \frac{1}{|A|} \sum_{a_i \in A}\mem{set}[a_i \in B] \text{,} \label{eq:exact-R}\\
    P &= \frac{1}{|B|} \sum_{b_i \in B}\mem{set}[b_i \in A]\text{,} \label{eq:exact-P}
\end{align}

\paragraph{Basic set operations}
We need three basic set operations: \emph{intersection} ($\set{A}$ \blue{$\cap$} $\set{B}$), \emph{union} ($\set{A}$ \blue{$\cup$} $\set{B}$), and \emph{complement} (\blue{$\overline{\set{A}}$}). 
They allow us to represent various sets using such different combinations as $\set{Color} \cap \set{Fruit}$.

\begin{table*}[t]
\centering
\begin{tabular}{cccccc}
\toprule
\multicolumn{2}{c}{\textbf{Symbolic Set Representation}} & & \multicolumn{3}{c}{\textbf{Subspace-based Set Representation}}\\
\midrule
\midrule
\hspace{-2mm}
\begin{tabular}{c}
Element\\
$\w{king}$
\end{tabular} 
& 
\hspace{-3mm}
\begin{minipage}{9mm}
\centering
\scalebox{0.025}{\includegraphics{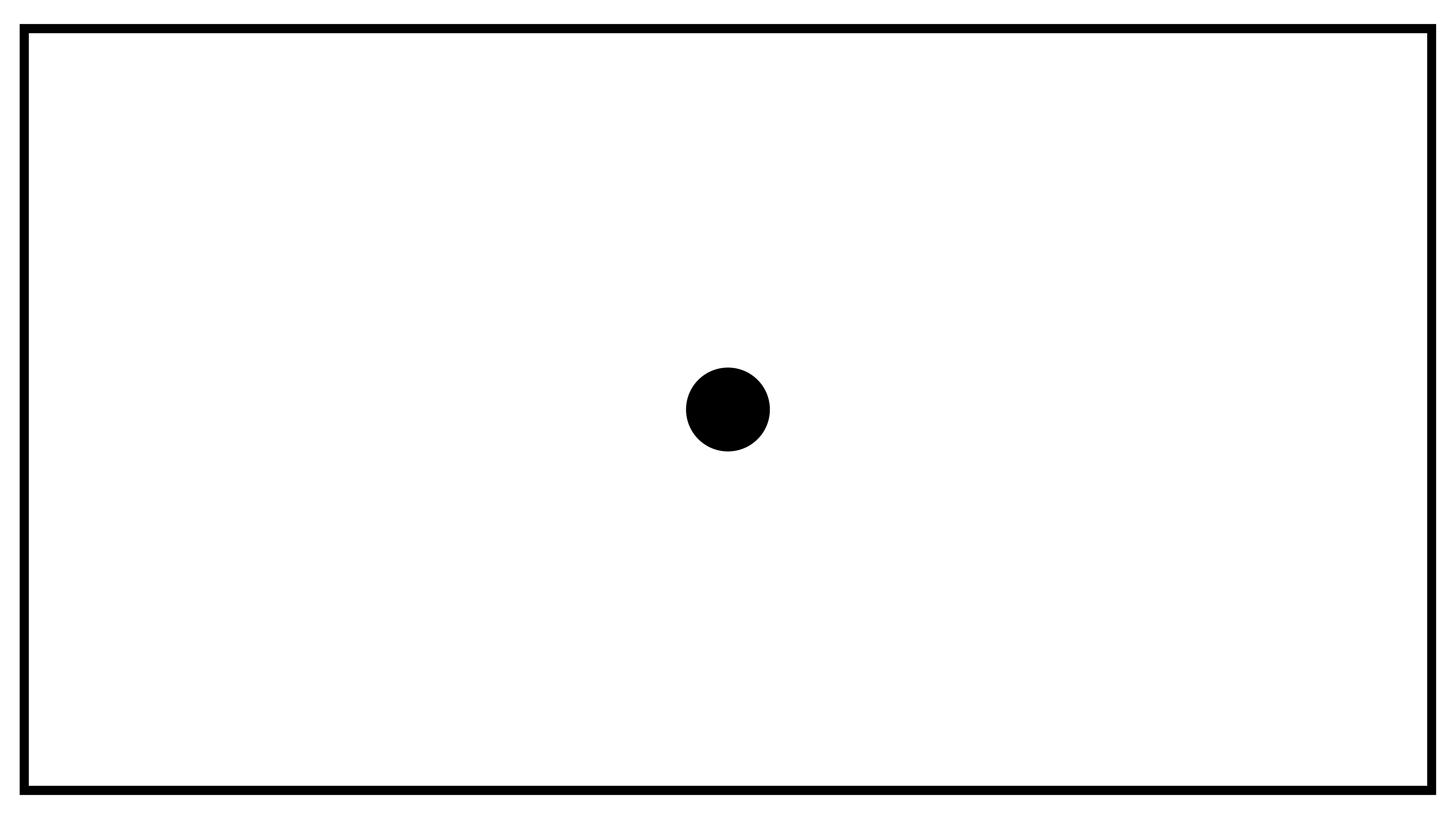}}
\end{minipage}
& &
\hspace{3mm}
\begin{tabular}{c}
Vector\\
$\v{king}$
\end{tabular} 
&
\begin{minipage}{12mm}
\scalebox{0.1}{\includegraphics{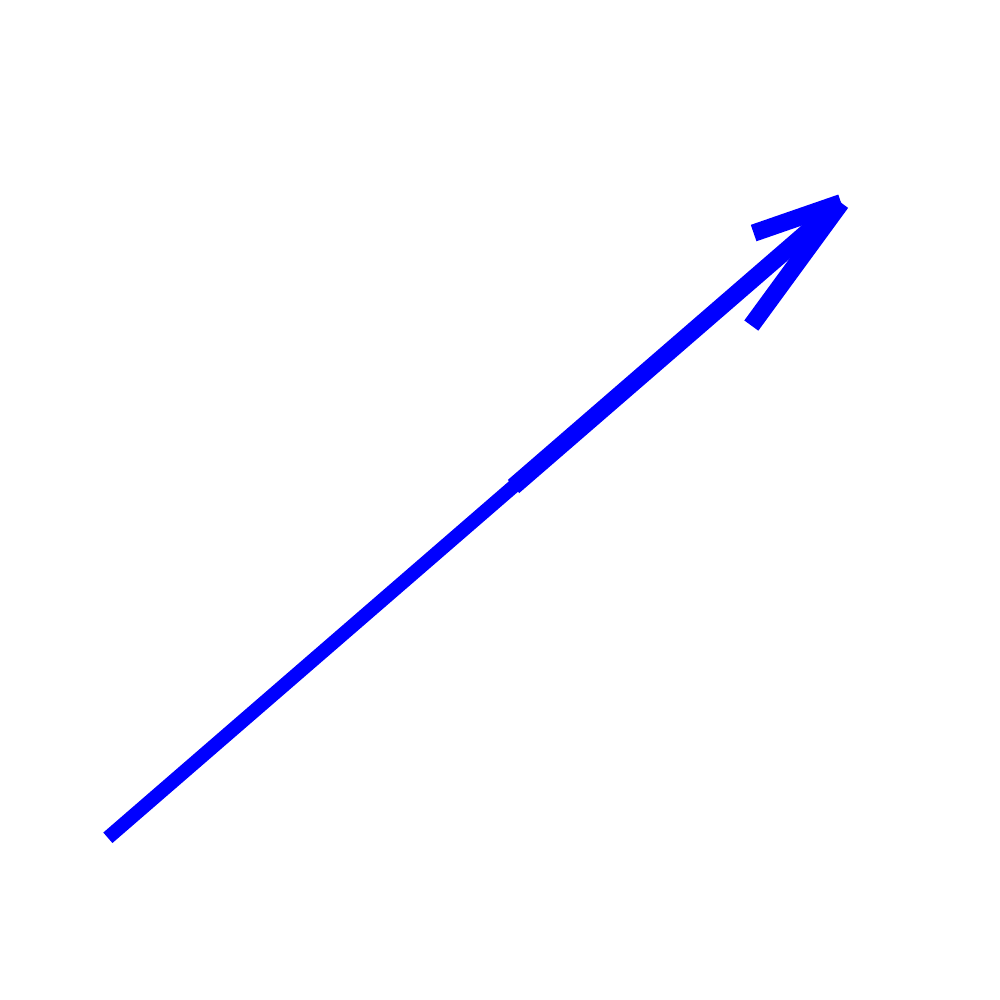}} 
\end{minipage}
&
\\
\midrule
\hspace{-2mm}
\begin{tabular}{c}
Set\\
$\set{Male} = \{\w{king}, \w{man}, \dots\}$
\end{tabular} 
& 
\hspace{-3mm}
\begin{minipage}{9mm}
\centering
\scalebox{0.025}{\includegraphics{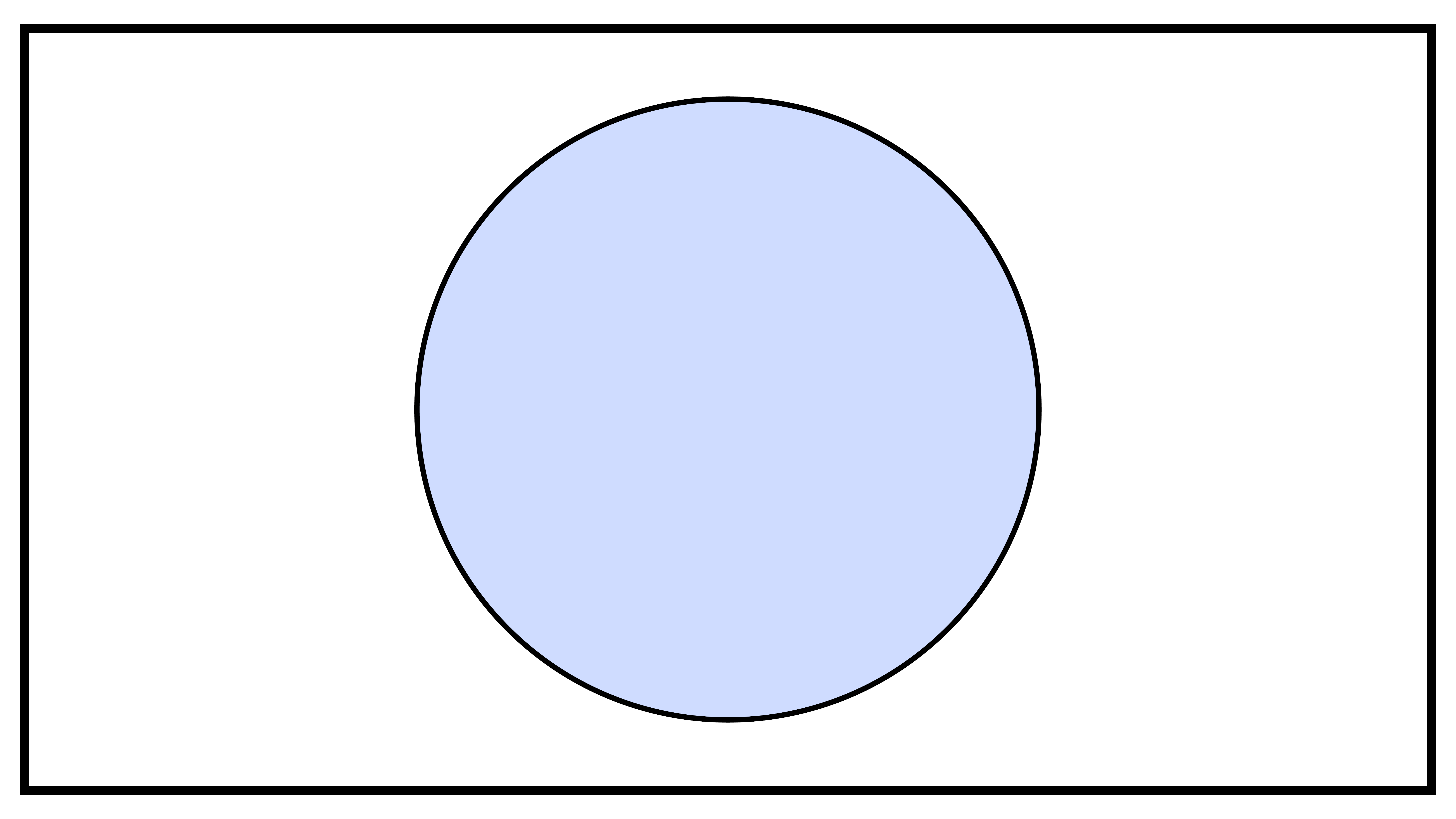}}
\end{minipage}
& &
\hspace{3mm}
\begin{tabular}{c}
Subspace\\
$\Subs{Male} = \mathrm{span}(\v{king}, \v{man}, \dots)$
\end{tabular} 
&
\hspace{-4mm}
\begin{minipage}{12mm}
\scalebox{0.18}{\includegraphics{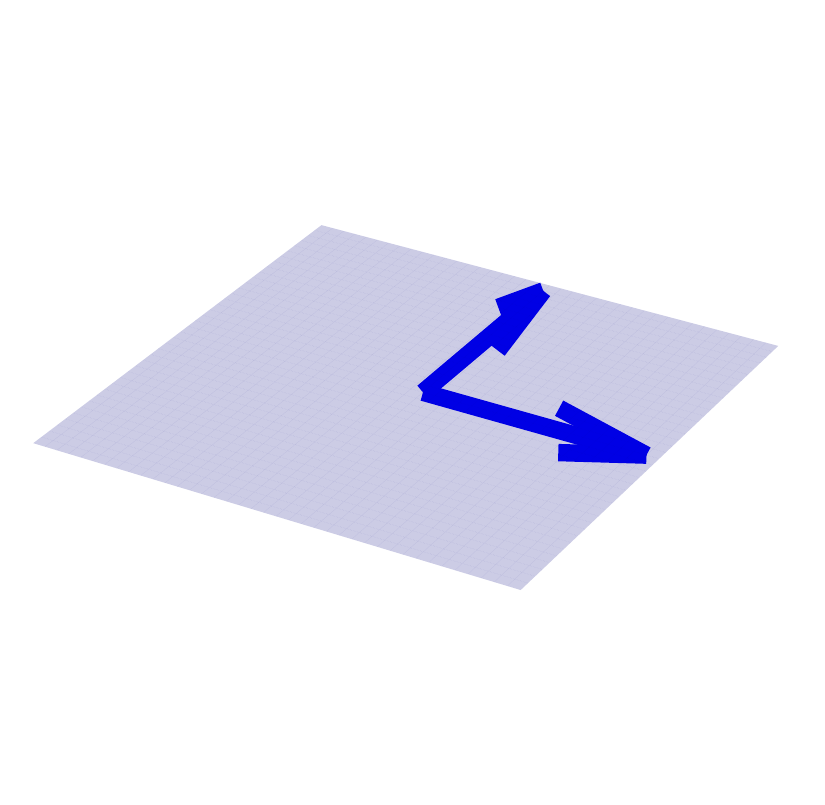}} 
\end{minipage}
&
\\ 
\midrule
\hspace{-2mm}
\begin{tabular}{c}
Complement\\
$\overline{\set{Male}}$
\end{tabular} 
& 
\hspace{-3mm}
\begin{minipage}{9mm}
\centering
\scalebox{0.025}{\includegraphics{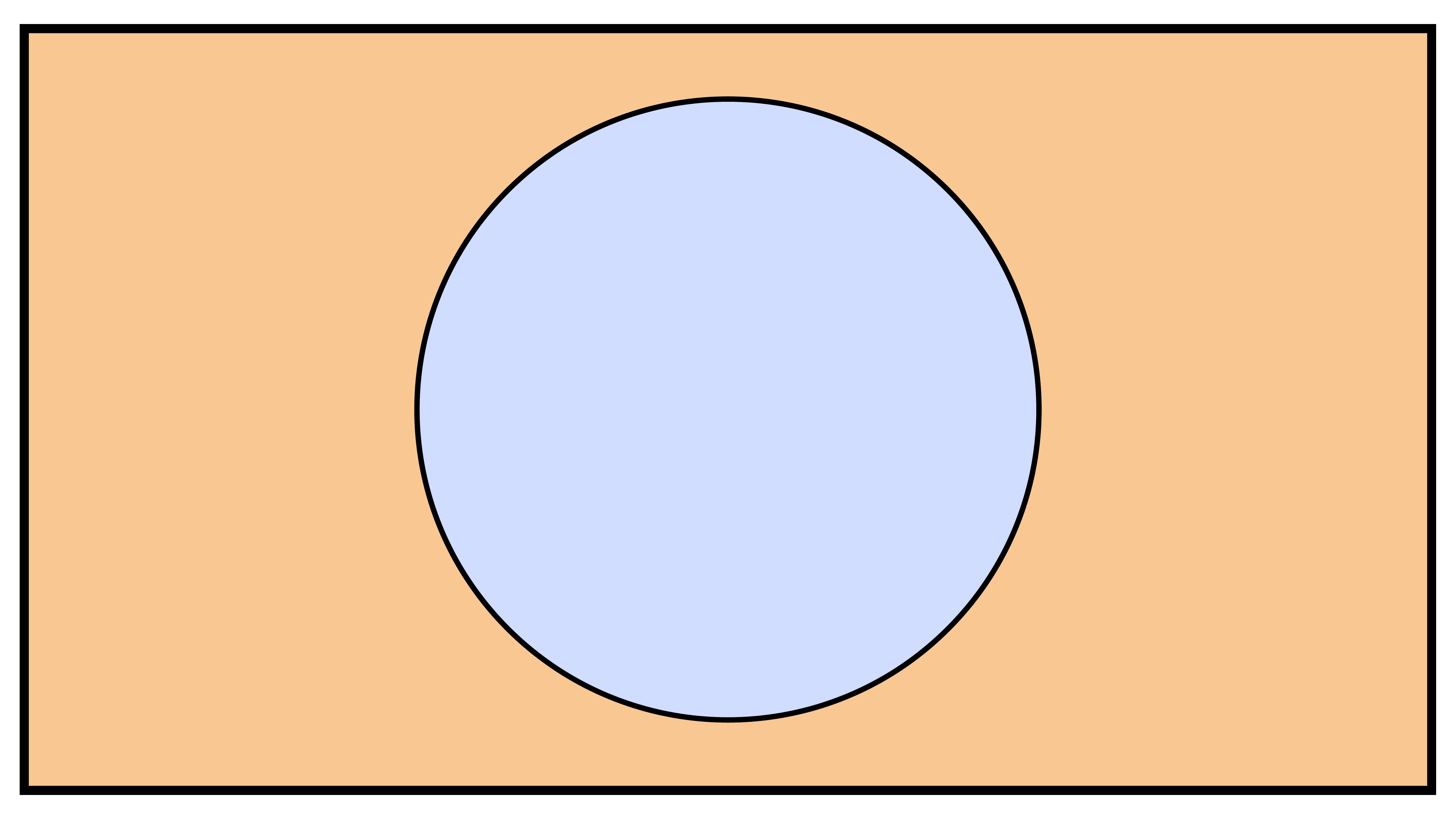}} 
\end{minipage}
& &
\hspace{3mm}
\begin{tabular}{c}
Orthogonal complement\\
$(\Subs{Male})^{\bot}$
\end{tabular} 
&
\hspace{-4mm}
\begin{minipage}{12mm}
\scalebox{0.14}{\includegraphics{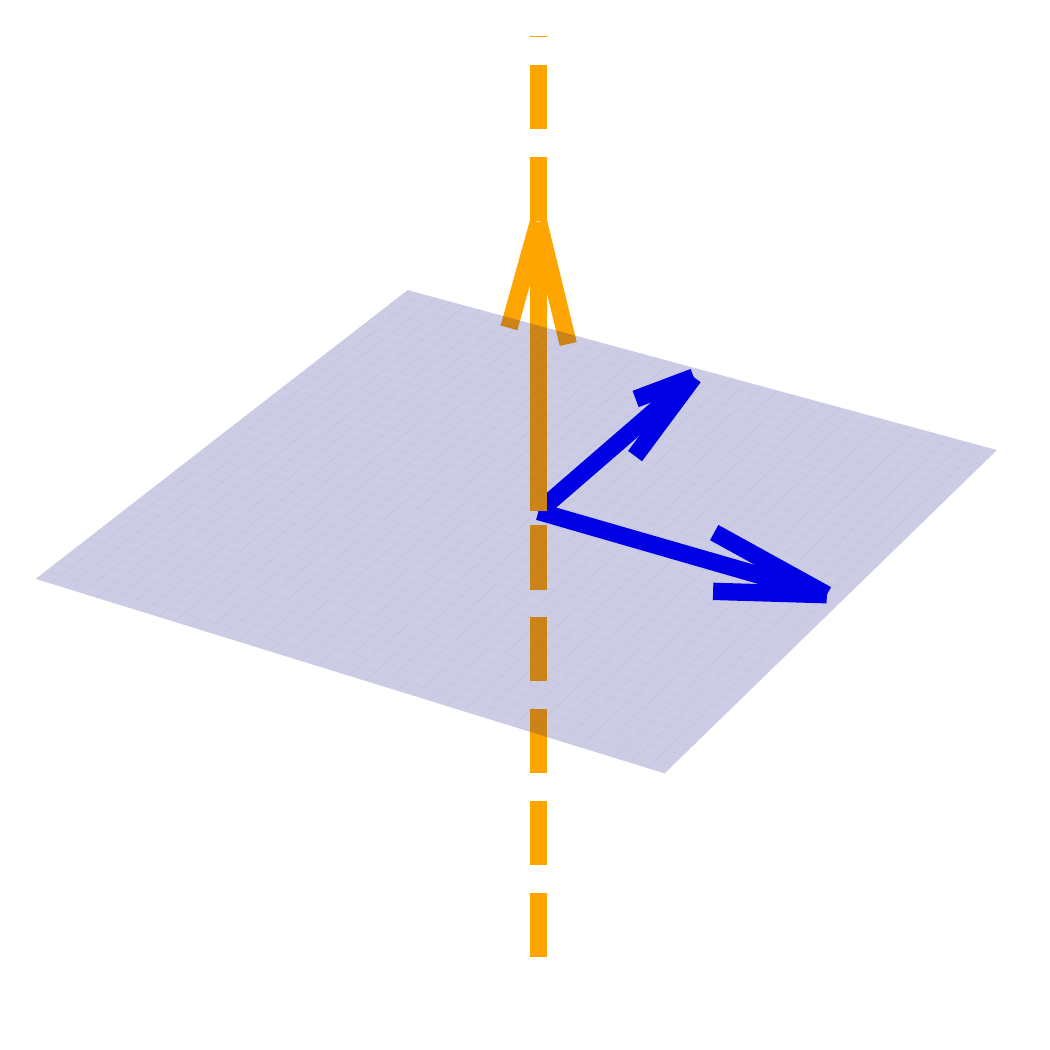}} 
\end{minipage}
&
\\
\midrule
\hspace{-2mm}
\begin{tabular}{c}
Union\\
$\set{Male} \cup \set{Female}$
\end{tabular} 
& 
\hspace{-3mm}
\begin{minipage}{9mm}
\centering
\scalebox{0.025}{\includegraphics{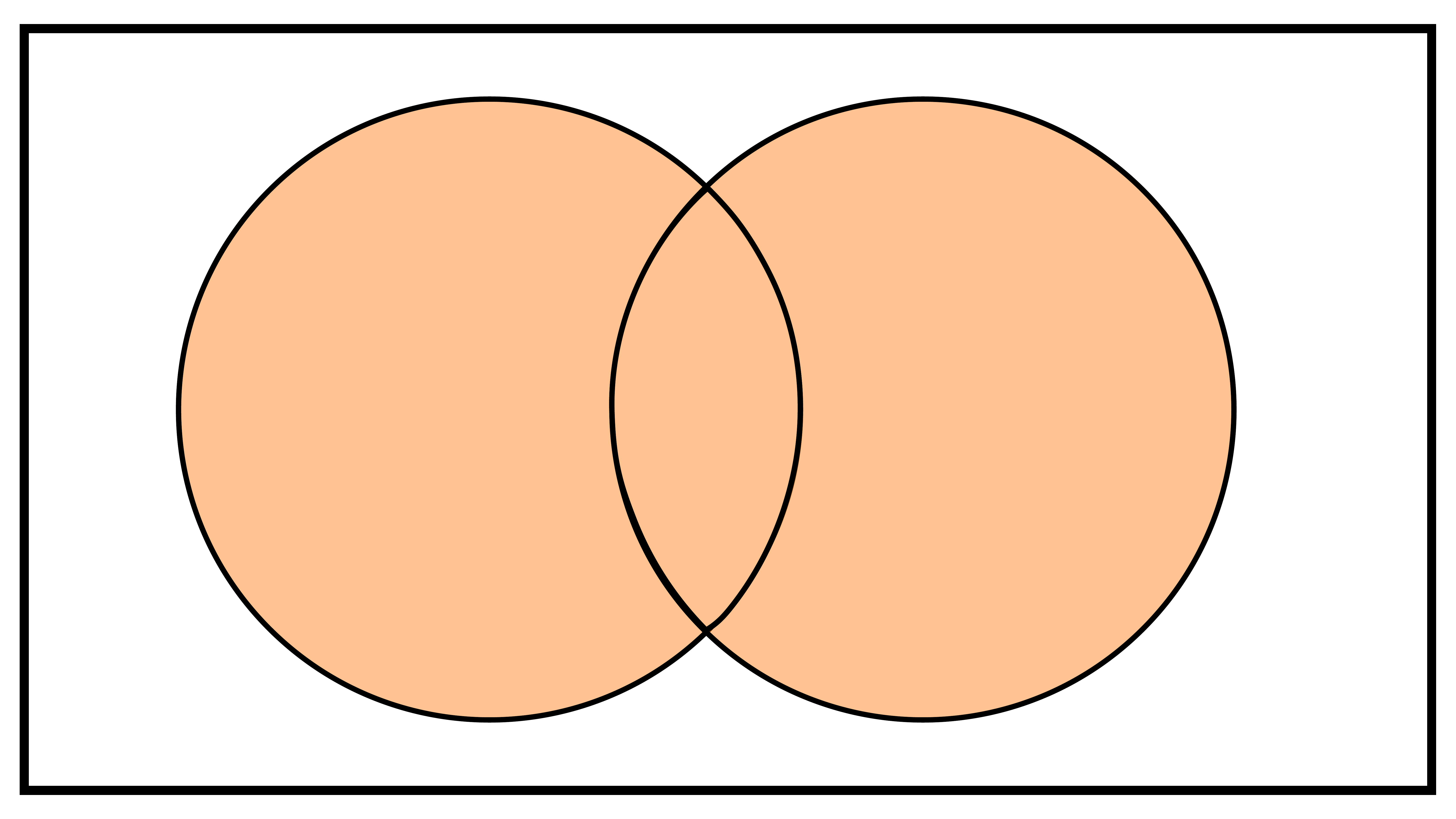}}
\end{minipage}
& &
\hspace{3mm}
\begin{tabular}{c}
Sum space\\
$\Subs{Male} + \Subs{Female}$
\end{tabular} 
&
\hspace{-4mm}
\begin{minipage}{12mm}
\scalebox{0.065}{\includegraphics{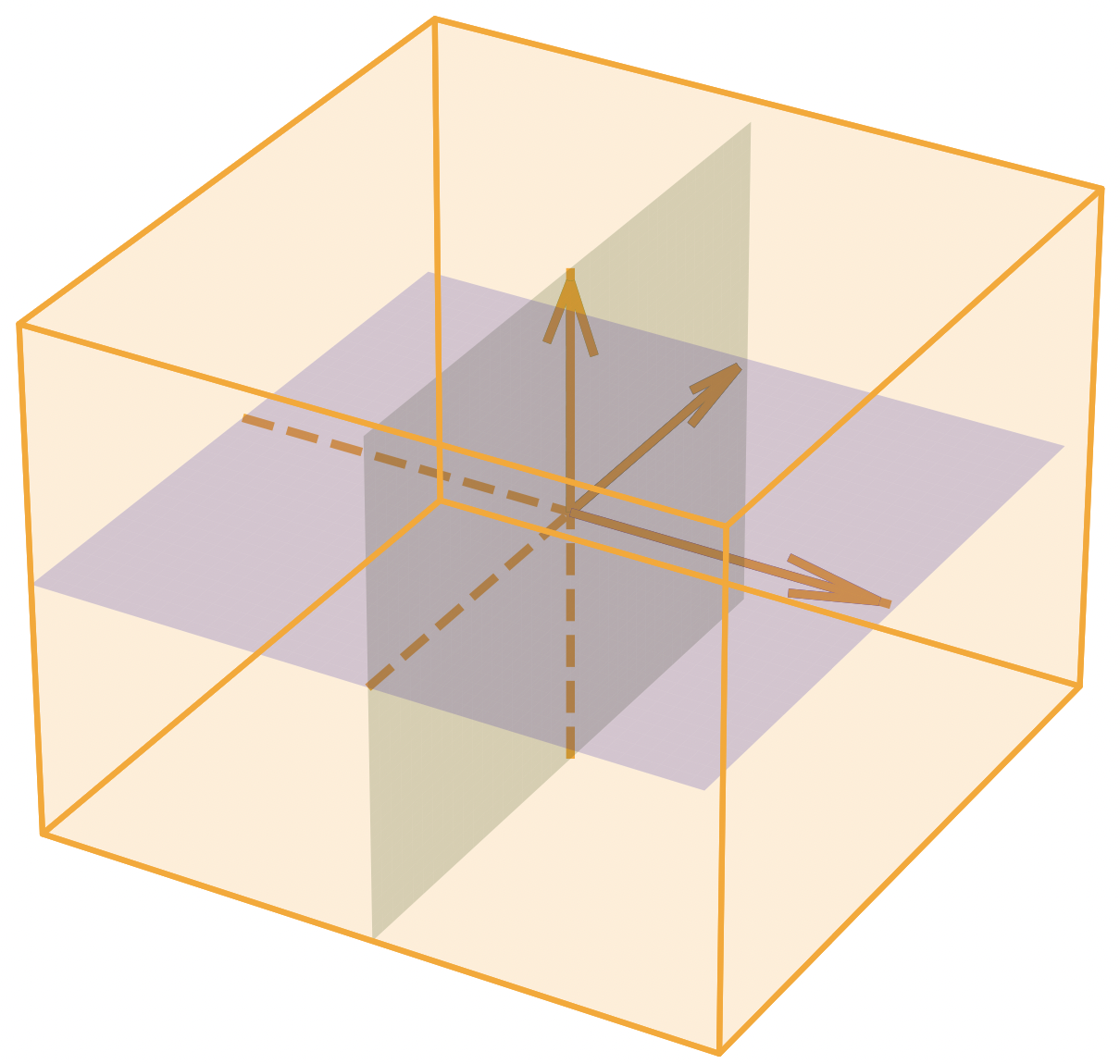}}
\end{minipage}
&
\\
\midrule
\hspace{-2mm}
\begin{tabular}{c}
Intersection\\
$\set{Color} \cap \set{Fruit}$
\end{tabular} 
& 
\hspace{-3mm}
\begin{minipage}{9mm}
\centering
\scalebox{0.025}{\includegraphics{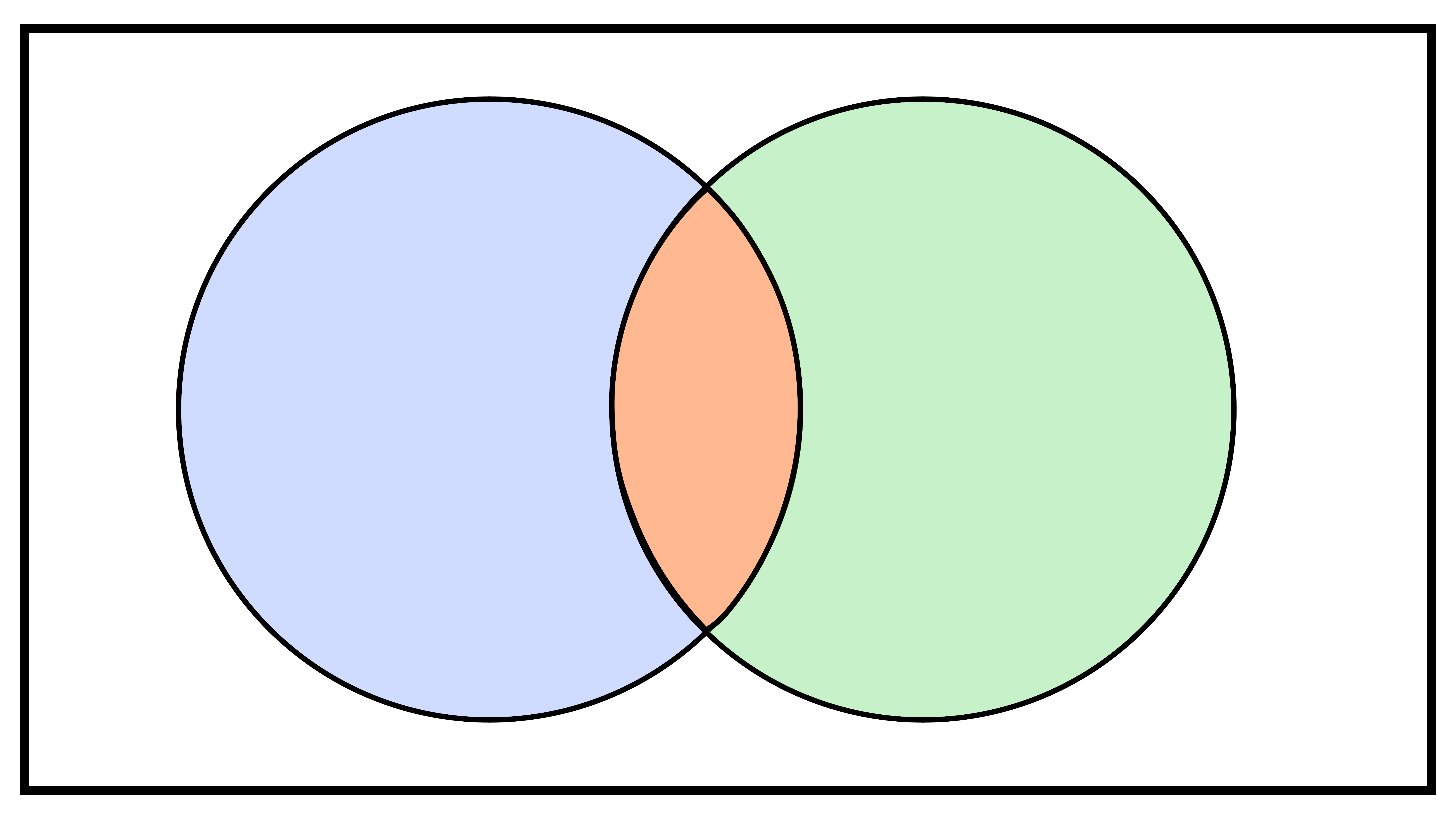}}
\end{minipage}
& &
\hspace{3mm}
\begin{tabular}{c}
Intersection\\
$\Subs{Color} \cap \Subs{Fruit}$
\end{tabular} 
&
\hspace{-4mm}
\begin{minipage}{12mm}
\scalebox{0.19}{\includegraphics{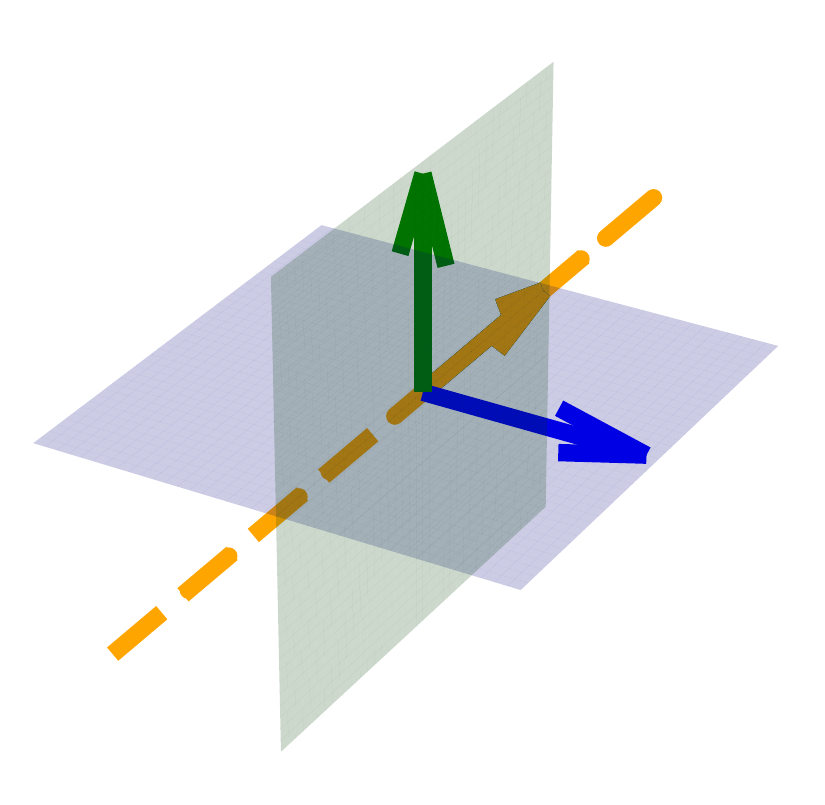}} 
\end{minipage}
&
\\
\midrule
\hspace{-2mm}
\begin{tabular}{c}
Set membership \\
$\mem{set}[\w{boy} \in \set{Male}]$
\end{tabular} 
&
\hspace{-3mm}
\begin{minipage}{9mm}
\centering
\scalebox{0.025}{\includegraphics{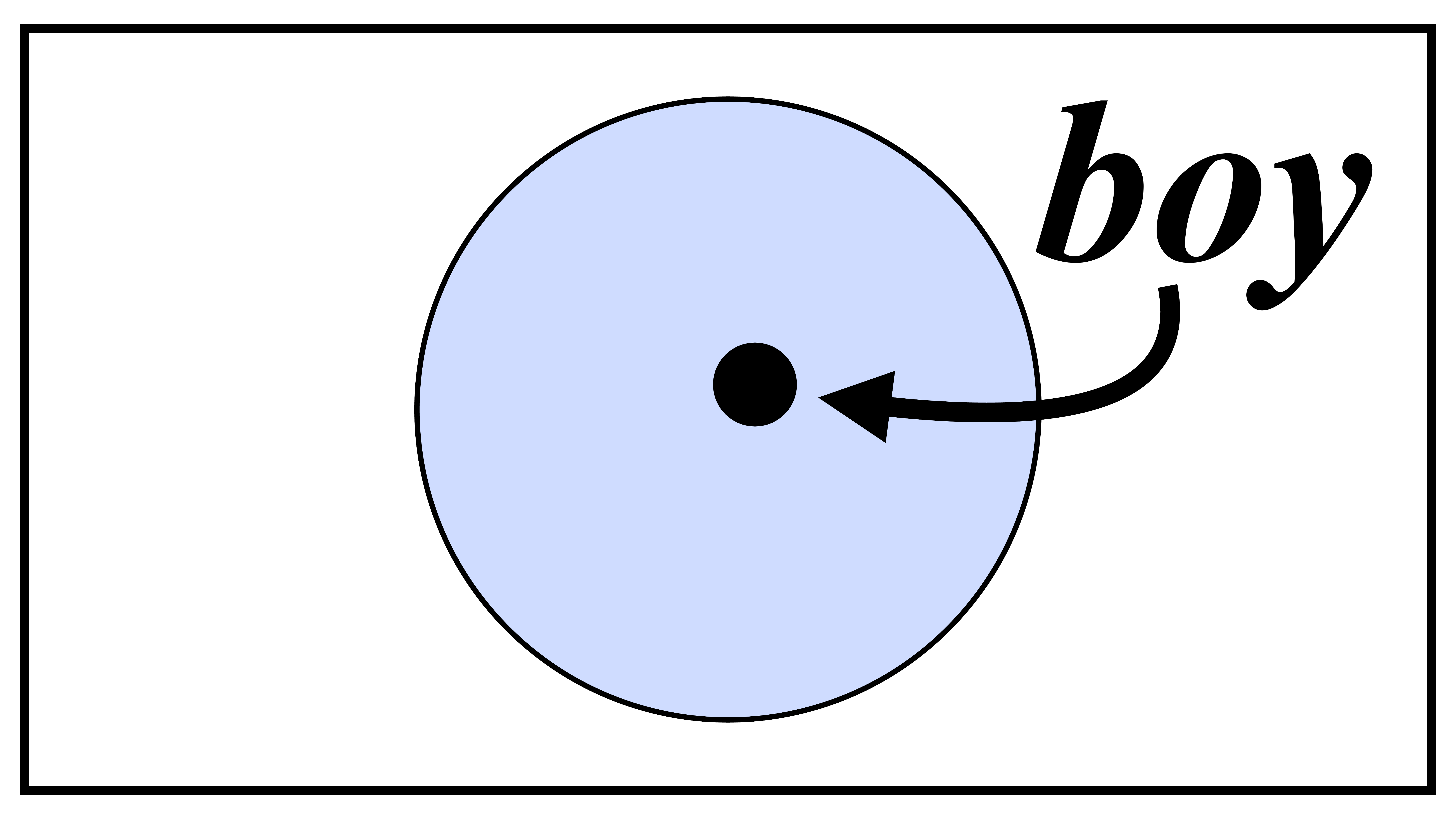}}
\end{minipage}
& & 
\hspace{3mm}
\begin{tabular}{c}
Subspace indicator function \\
$\softin(\v{boy}, \Subs{Male})$
\end{tabular} 
&
\hspace{-3.8mm}
\begin{minipage}{12mm}
\scalebox{0.067}{\includegraphics{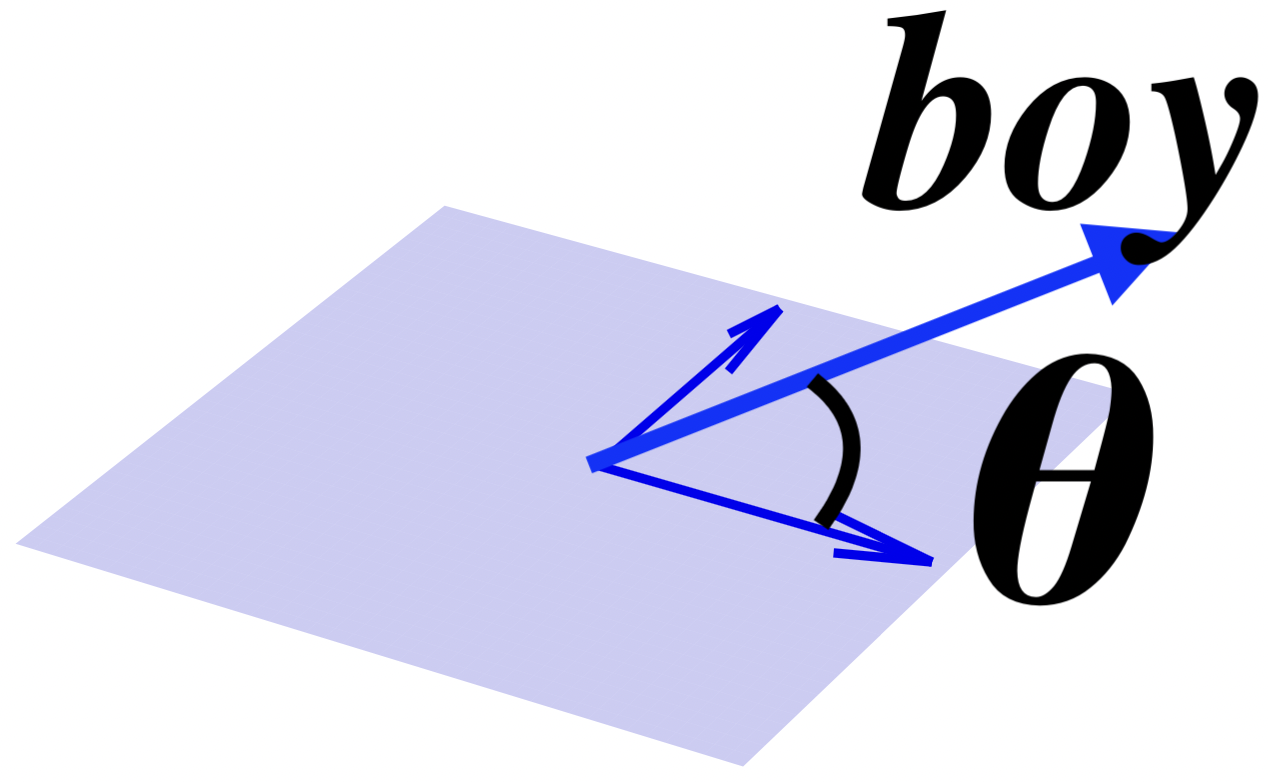}}
\end{minipage}
&
\\
\midrule
Set similarity & & & SubspaceBERTScore \\
\hspace{-2mm}
\begin{tabular}{c}
Recall \\
\end{tabular} 
&
\hspace{-9mm}
\begin{minipage}{12mm}
\scalebox{0.04}{\includegraphics{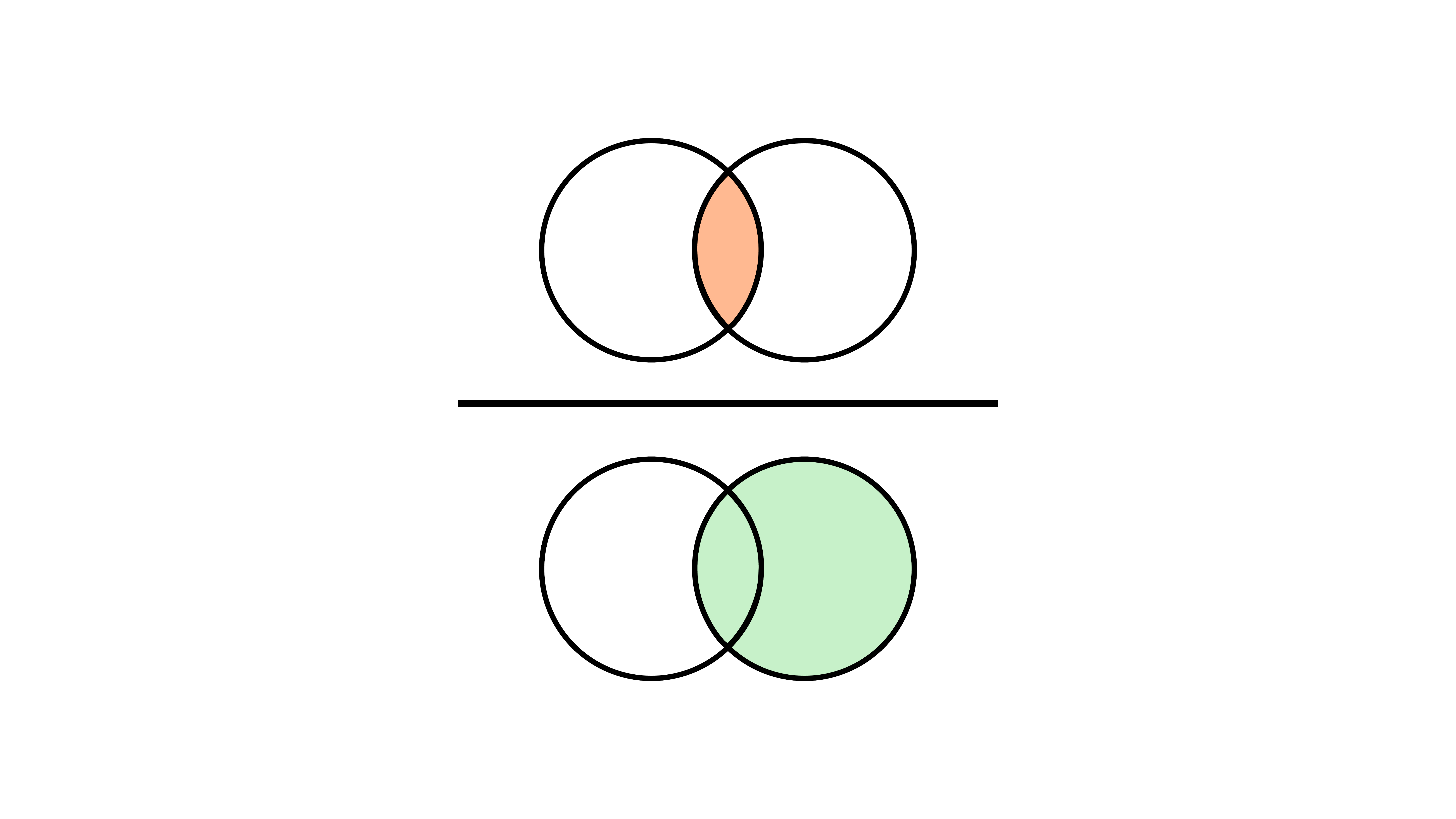}}
\end{minipage}
& &
\hspace{1mm}
\begin{tabular}{c}
$R_{\mathrm{subspace}}$
\end{tabular} 
&
\hspace{-17mm}
\begin{minipage}{12mm}
\scalebox{0.04}{\includegraphics{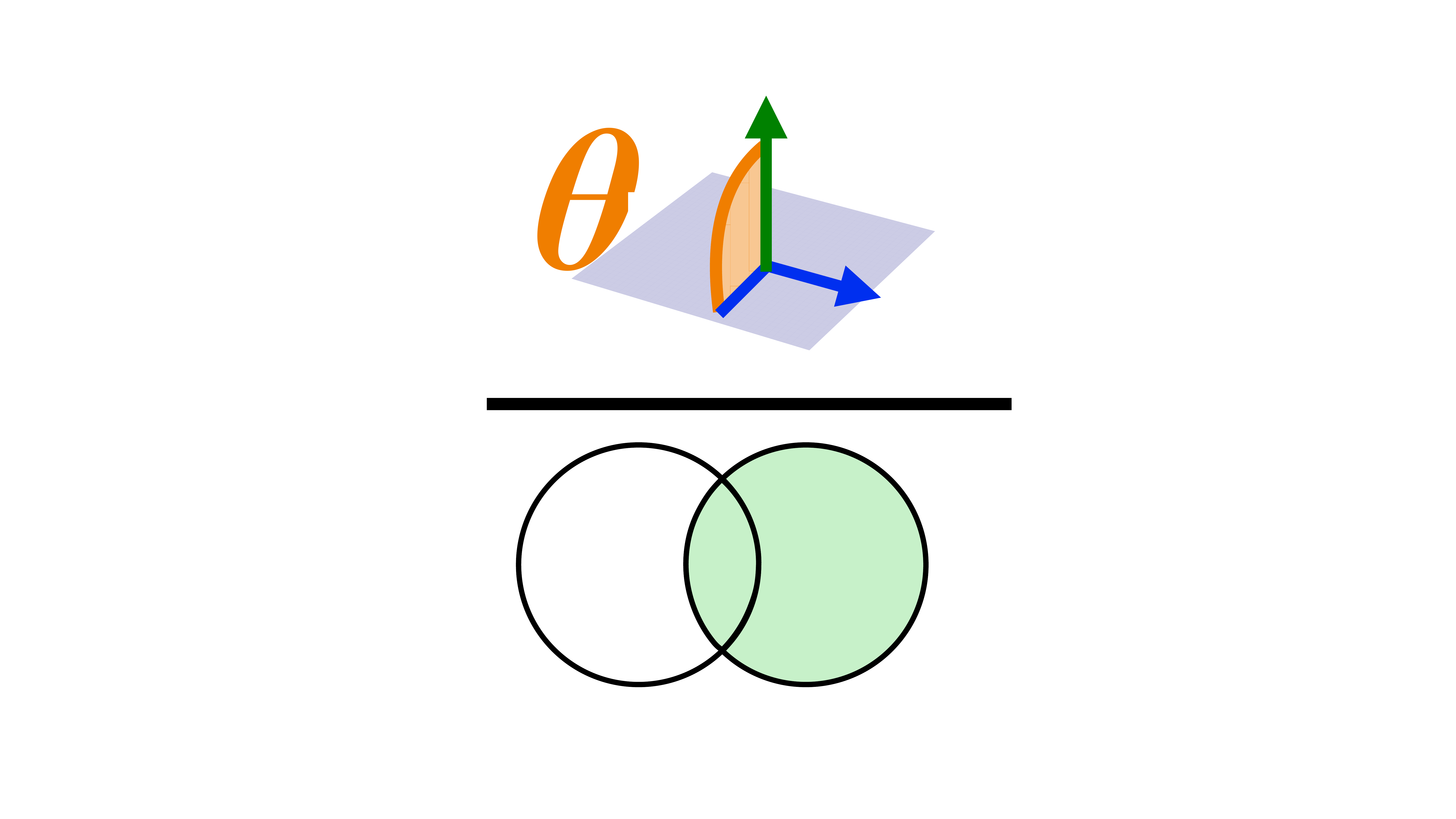}}
\end{minipage}
&
\\
\hspace{-2mm}
\begin{tabular}{c}
Precision \\
\end{tabular} 
&
\hspace{-9mm}
\begin{minipage}{12mm}
\scalebox{0.04}{\includegraphics{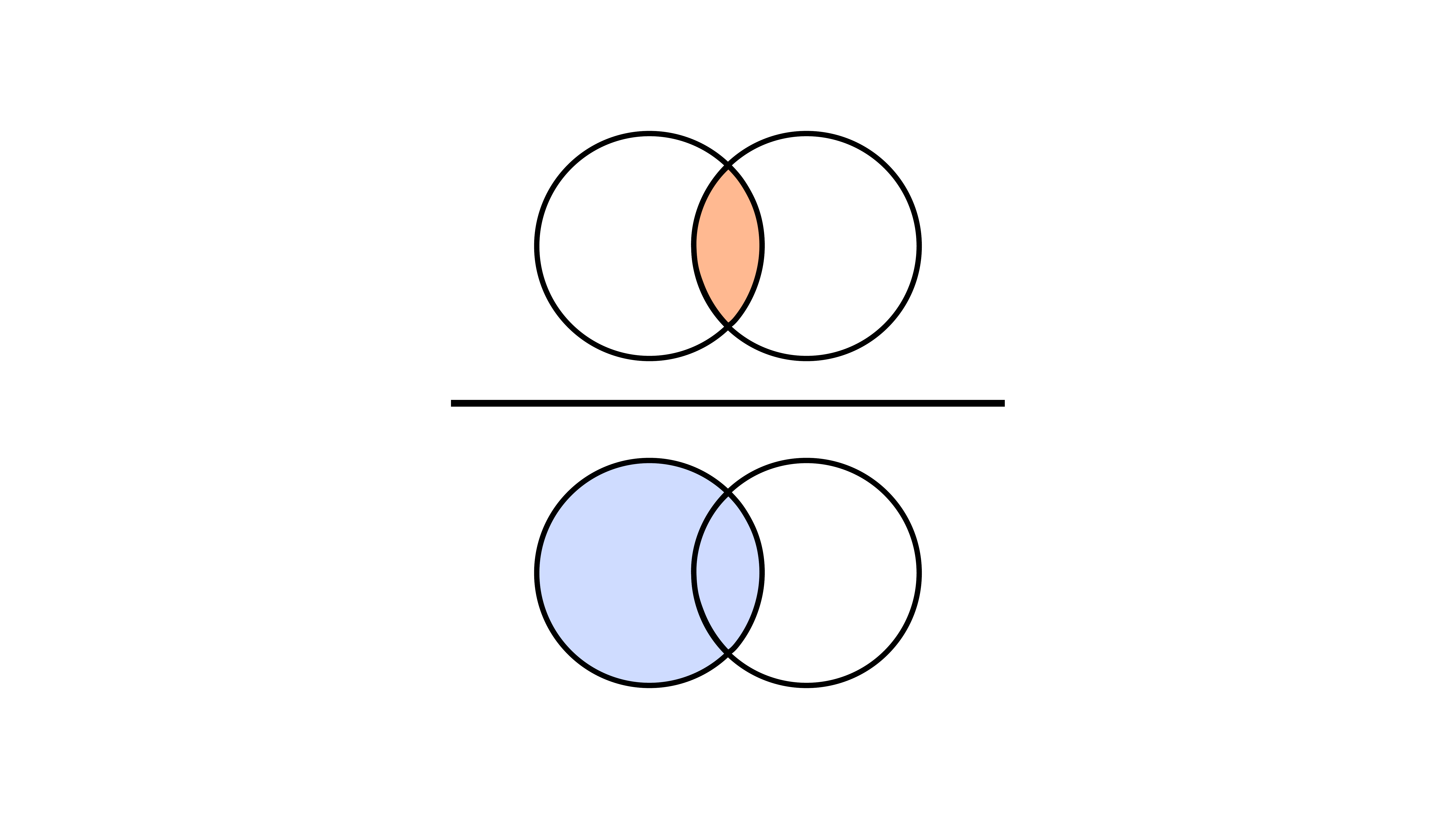}}
\end{minipage}
& &
\hspace{1mm}
\begin{tabular}{c}
$P_{\mathrm{subspace}}$
\end{tabular} 
&
\hspace{-17mm}
\begin{minipage}{12mm}
\scalebox{0.04}{\includegraphics{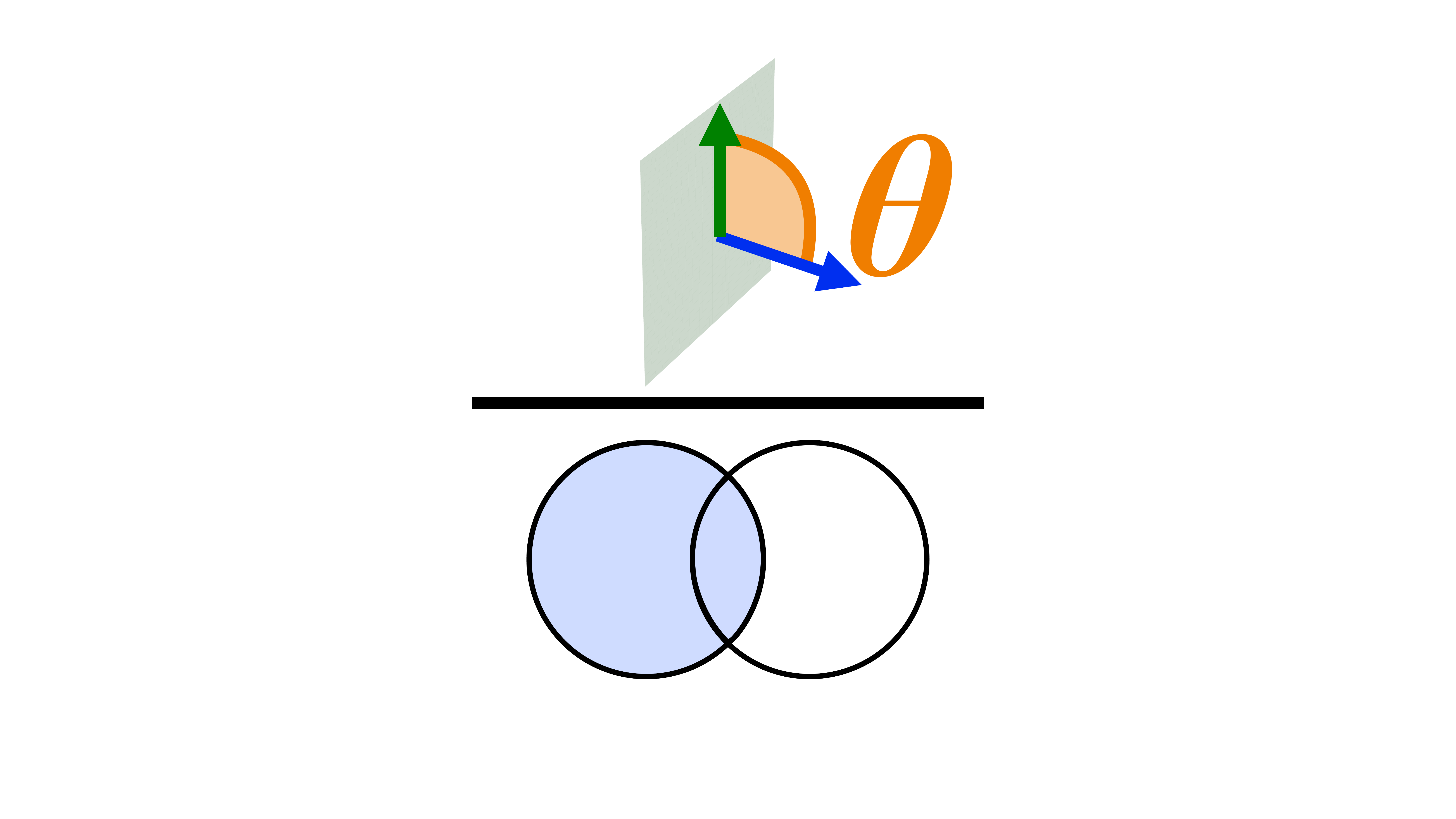}}
\end{minipage}
&
\\
\hspace{-2mm}
\begin{tabular}{c}
F-score
\end{tabular} 
&
\hspace{-21mm}
\begin{minipage}{12mm}
\scalebox{0.05}{\includegraphics{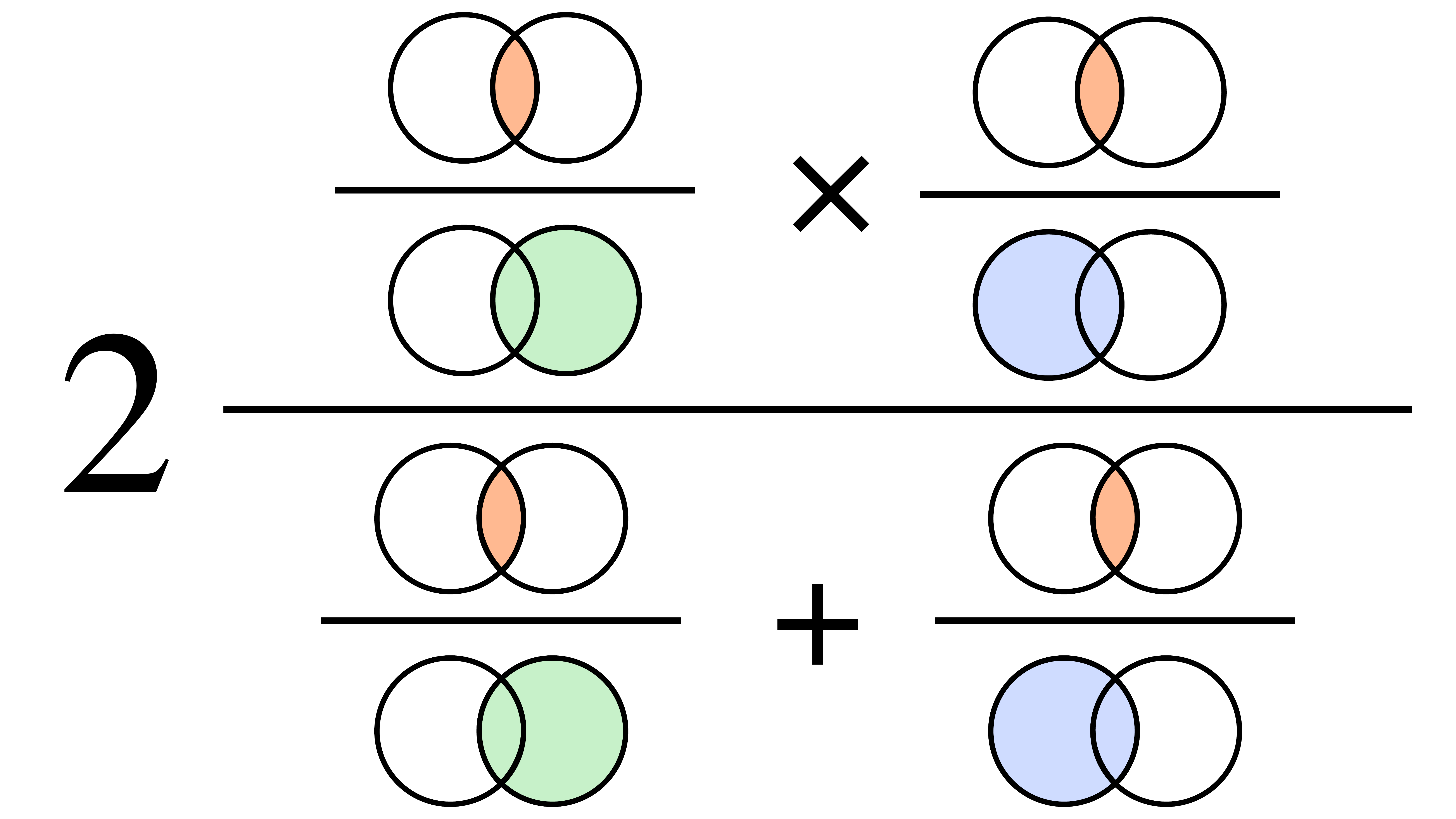}}
\end{minipage}
& &
\hspace{1mm}
\begin{tabular}{c}
$F_{\mathrm{subspace}}$
\end{tabular} 
&
\hspace{-29mm}
\begin{minipage}{12mm}
\scalebox{0.05}{\includegraphics{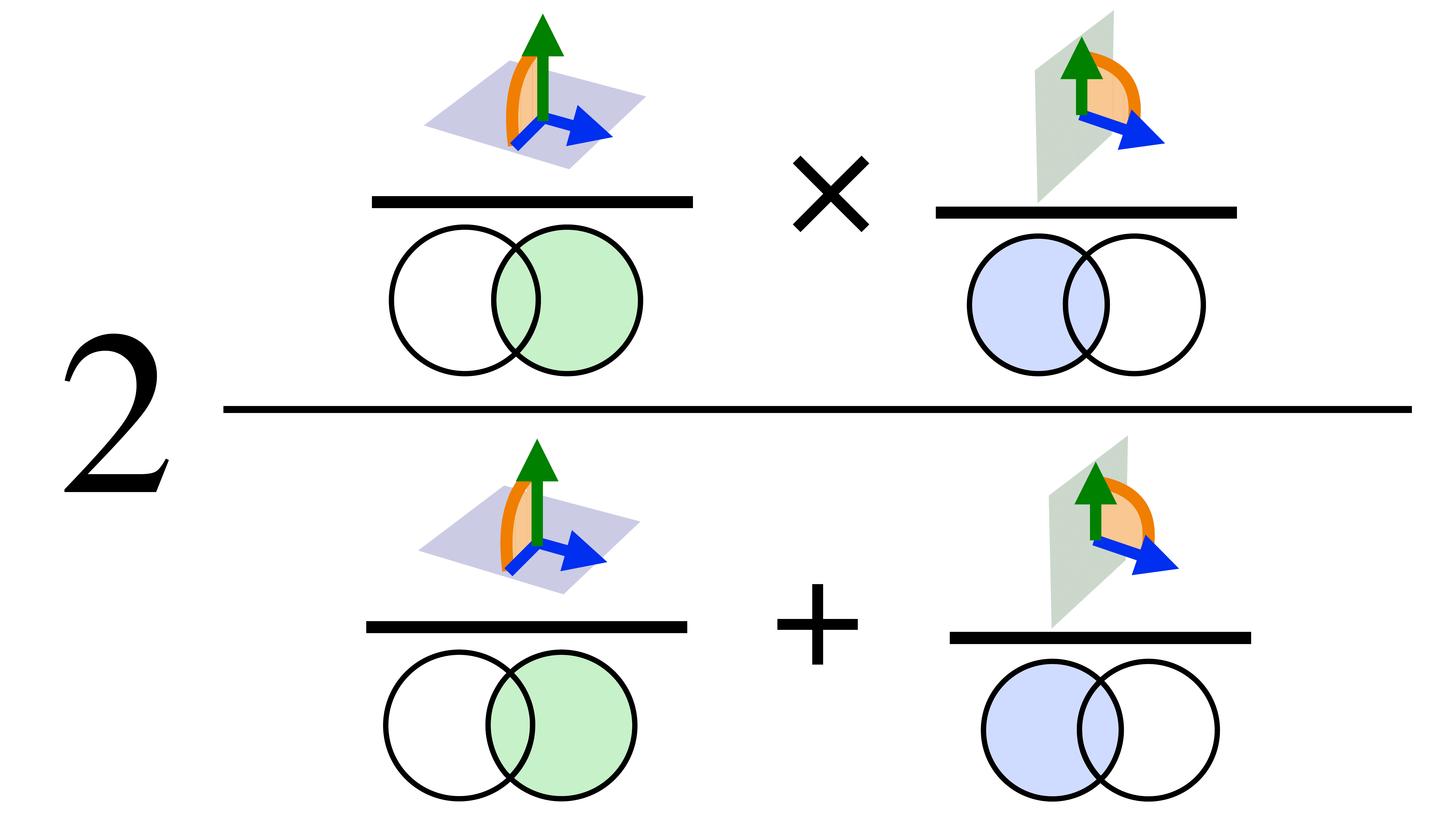}}
\end{minipage}
&
\\
\bottomrule
\end{tabular}
\caption{
Correspondence between symbolic set representations and subspace-based set representations:
We demonstrate that union, intersection, and complement, which are formulated in quantum logic, and our new formulations of set membership and word set similarity hold in pre-trained word embedding space.
}
\label{tab:subspace}
\end{table*}

\section{Subspace-based Set Representations}
\label{sec:quantum}
We propose the representations of a word set and set operations based on \emph{quantum logic} \cite{birkhoff1936logic}.
They hold advantages of geometric properties in an embedding space, and the set operations are guaranteed to hold for the laws of a set defined in quantum logic.

\subsection{Quantum logic}
While word embedding represents a word's meaning as a vector in linear space, quantum mechanics similarly represents a quantum state as a vector in linear space.
These two intuitively different fields are very close to each other in terms of the representation and the operation of information.

Quantum logic \cite{birkhoff1936logic} theory describes quantum mechanical phenomena.
Intuitively, it is a framework for \emph{set operations in a vector space}. 
In quantum logic, a set of vectors is represented as a linear subspace in a Hilbert space, and such set operations as union, intersection, and complement are defined as operations on subspaces.
Quantum logic, which employs a complete orthomodular lattice as its system of truth values, guarantees to hold various set operations, such as De Morgan's laws $\overline{(\set{A} \cap \set{B})} = \overline{\set{A}} \cup \overline{\set{B}}$ and $\overline{(\set{A} \cup \set{B})} = \overline{\set{A}} \cap \overline{\set{B}}$, idempotent law: $\set{A} \cap \set{A} = \set{A}$, and double complement: $\overline{\overline{\set{A}}} = \set{A}$.

\subsection{Set Operations in an Embedding Space}
The representations of an element, a set, and such set operations as union, intersection, and complement in quantum logic can be applied directly in a word embedding space because it is a Euclidean space and therefore also a Hilbert space.
However, since set similarity and set membership for a word embedding space are still missing in quantum logic, we propose a novel formulation of those operations using subspace-based representations, which is consistent with quantum logic.
The correspondence between symbolic and subspace-based set operations is shown in \autoref{tab:subspace}.

\paragraph{Set and elements}
\begin{algorithm}[t]
    \caption{Computing basis of a subspace}
    \label{alg:subspace}
    \begin{algorithmic}
    \Require $\{\v{}^{(1)}, \ldots, \v{}^{(k)}\} \subseteq \mathbb{R}^{1 \times d}$: Word embeddings to span subspace $\Subs{A}$
    \Ensure $\ms{A} \in \mathbb{R}^{r \times d}$: Bases of $\Subs{A}$
    \State $\m{A} \in \mathbb{R}^{k\times d} \leftarrow \textsc{stack\_rows}(\v{}^{(1)}, \ldots, \v{}^{(k)})$
    \State $\ms{A} \in \mathbb{R}^{r \times d} \leftarrow (\textsc{ortho\_normal}(\m{A}^{\top}))^{\top}$
    \Comment{Orthonormalize the bases. $r$ is the rank of $\m{A}$}    
    \State \Return $\ms{A}$
    \end{algorithmic}
\end{algorithm}
Let $\mathbb{R}^n$ be a $n$-dimensional embedding space (Euclidean space), let $\set{A} = \{\w{w}_1, \w{w}_2, \dots\}$ be a set of words, and let $\v{w} \in \mathbb{R}^n$ be a word (token) vector corresponding to $\w{w}$.
As discussed in \S\ref{sec:problem}, we first formulate the representation of a word and a word set.
In quantum logic, an element is represented by a vector, and a set is represented by a subspace spanned by the vectors corresponding to its elements.
Here we assume an element, i.e., word $\w{w}$, is represented by vector $\v{w}$, and a word set is represented by linear subspace $\Subs{A} \subset \mathbb{R}^n$ spanned by word vectors:
\begin{equation}
\begin{split}
    \Subs{A} 
    \coloneqq \mathrm{span}(\m{A}) 
    \coloneqq \mathrm{span}(\vec{a}_1, \vec{a}_2, \dots)
    \text{.}
    \label{eq:span}
\end{split}
\end{equation}
Hereinafter we simply refer to \emph{linear subspace} as \emph{subspace}.
Algorithm \ref{alg:subspace} is the pseudocode for computing the basis of the subspace.

\paragraph{Basic set operations}
The \emph{complement} of set $\set{A}$, denoted by $\overline{\set{A}}$, is represented by the orthogonal complement of subspace $\Subs{A}$: 
\begin{equation}
\begin{split}
    \Subs{\overline{A}} 
    \coloneqq (\Subs{A})^{\bot}
    = \{\vec{v} \mid \exists\vec{a} \in \Subs{A}, \vec{v} \cdot \vec{a} = 0\}
    \text{.}
\end{split}
\end{equation}

The \emph{union} of two sets, $\set{A}$ and $\set{B}$, denoted by $\set{A} \cup \set{B}$, is represented by the sum space of two subspaces, $\Subs{A}$ and $\Subs{B}$:
\begin{equation}
\begin{split}
    \Subs{A \cup B} 
    \!\coloneqq \Subs{A} + \Subs{B} 
    \!=\! \{\vec{a} + \vec{b} \!\mid\! \vec{a} \!\in\! \Subs{A},\vec{b} \!\in\! \Subs{B}\}\text{.}\!\!\!
\end{split}
\end{equation}

The \emph{intersection} of two sets, $\set{A}$ and $\set{B}$, denoted by $\set{A} \cap \set{B}$, is represented by the intersection of two subspaces, $\Subs{A}$ and $\Subs{B}$:
\begin{equation}
\begin{split}
    \Subs{A \cap B} 
    \coloneqq \Subs{A} \cap \Subs{B} 
    = \{\vec{v} \mid \vec{v} \in \Subs{A}, \vec{v} \in \Subs{B}\}
    \text{.}
\end{split}
\end{equation}

The basis of the intersection can be computed based on singular value decomposition (SVD). The bases are the vectors shared by the two subspaces.

\paragraph{Hard membership}
The set membership in the embedding space (e.g., $\w{boy} \in \set{Male}$) can be represented by the inclusion of a vector into a subspace (e.g., $\v{boy} \in \Subs{Male}$) and
given by the following indicator function:
\begin{equation}
    \hardin(\v{}, \Subs{A}) \coloneqq 
    \begin{cases}
        1 \quad(\v{} \in \Subs{A}), \\
        0 \quad(\v{} \notin \Subs{A}). \\
      \end{cases}
\label{eq:hard_membership}
\end{equation}
However, this binary decision fails to exploit the geometric properties of the word embedding space regarding semantic similarity.
Suppose we quantify the degree of membership of word $\w{boy}$ for word set $\set{Male}$ consisting of many masculine nouns other than $\w{boy}$.
Even if $\v{boy}$ is located very close to $\Subs{Male}$ due to its semantic similarity to masculine nouns, $\hardin(\v{boy}, \Subs{Male})$ must return $0$ because $\v{boy}$ must not be located \emph{exactly} on subspace $\Subs{Male}$ defined by \set{Male}.
It must return $1$ based on the masculine property of word $\w{boy}$.
Such \emph{hard} membership defined by Eq. \eqref{eq:hard_membership} is  incompatible with an embedding space.

\paragraph{Soft membership: Subspace indicator function}
Instead, we define another membership function called \emph{subspace indicator function} $\softin$ that returns continuous values from $0$ to $1$ depending on the following minimum angle between vector $\v{w}$ and subspace $\Subs{A}$ (the first canonical angle): 
\begin{equation}
\begin{split}
    \!\!\!
    \softin(\v{}, \Subs{A}) 
    & \coloneqq \max \left\{ \!\left.{\frac{|\vec{a} \cdot \v{}|}{\|\vec{a}\|\|\v{}\|}} \right|\vec{a}\in {\Subs{A}} \!\right\}\!\text{.}\!\!
    \label{eq:membership}
\end{split}
\end{equation}
This captures the degree of membership between a word and a word set, represented by the angle between a word vector and a subspace.
It is a natural extension of $\hardin$, i.e., $\softin$ returns $1$ when $\v{w} \in \Subs{A}$ and $0$ when $\v{w} \in \Subs{\overline{A}}$.

The key distinction of our subspace indicator function approach lies in its ability to leverage the comprehensive information encapsulated within pre-trained word embedding space. The subspace indicator function does not simply find the nearest individual word from the set. Instead, we consider the \emph{closeness} of the query word to the entire set as a whole, \textbf{by projecting the query word into the subspace spanned by the pre-trained embeddings} (as illustrated in the figure of the subspace indicator function function in \autoref{tab:subspace}). This way, we account not just for the individual word similarities, but also for the overall semantic coherence of the word set.
The detailed process for computing this subspace indicator function can be found in Algorithm \ref{alg:softind}.

\subsection{Set Similarity}
\label{sec:setsim}
\begin{figure*}[t]
    \centering
    \includegraphics[clip, width=14cm]{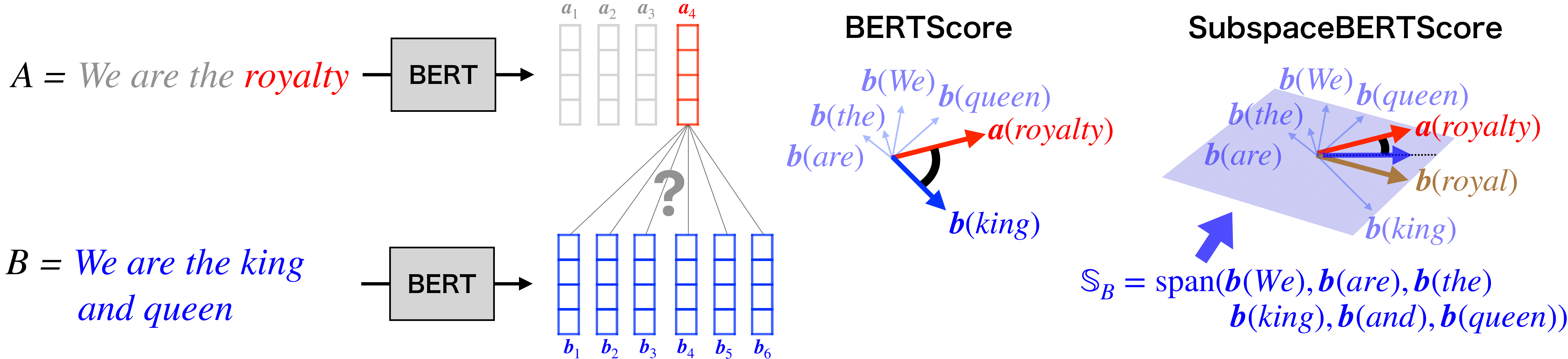}
    \caption{Comparison between the proposed SubspaceBERTScore and BERTScore. 
    We visualize the alignment process between the word \textit{royalty} and the words in the sentence $B$. 
    SubspaceBERTScore represents $B$ as the subspace $\Subs{B}$ and calculates the similarity (canonical angle) between $\Subs{B}$ and the \textit{royalty} vector ($\vec{a}_{4}$). 
    Our approach provides a ``softer'' alignment, capturing the overall semantic context of the sentence. 
    On the other hand, BERTScore adopts a ``harder'' alignment strategy, selecting only the word from the sentence with the maximum cosine similarity.}
    \label{fig:concept}
\end{figure*}

\paragraph{Limitation of symbolic set similarities}
Suppose we quantify the set similarity between $\set{A} = \{$\emph{A}, \emph{boy}, \emph{walks}, \emph{in}, \emph{this}, \emph{park}$\}$ and $\set{B} = \{$\emph{The}, \emph{kid}, \emph{runs}, \emph{in}, \emph{the}, \emph{square}$\}$,
which represent semantically similar sentences.
The challenge with traditional symbolic set similarities, such as recall, is that they primarily rely on the exact overlap of words between the sets. 
These semantically similar sentences share only one word $\{\w{in}\}$ between $\set{A}$ and $\set{B}$: $|\set{A} \cap \set{B}| = 1$. 
To address this shortcoming, it is essential to compute vector-based set similarity, such as BERTScore.

\paragraph{Three types of similarity in BERTScore}
BERTScore is a method that uses embeddings to approximately calculate $R$, $P$, and the $F$-score:
\begin{align}
    R_{\mathrm{BERT}} &= \frac{1}{|A|} \sum_{\vec{a}_i \in \m{A}}\mem{vectors}(\vec{a}_i, \m{B})\text{,} \\
    P_{\mathrm{BERT}} &= \frac{1}{|B|} \sum_{\vec{b}_i \in \m{B}}\mem{vectors}(\vec{b}_i, \m{A})\text{,} \\
    F_{\mathrm{BERT}} &= 2\frac{P_{\mathrm{BERT}} \cdot R_{\mathrm{BERT}}}{P_{\mathrm{BERT}} + R_{\mathrm{BERT}}}\text{,}       
\end{align}
where a sentence is represented as a set of token vectors $\m{A}$ and $\m{B}$.
$\mem{vectors}$ is the indicator function for vector sets.
Intuitively, this indicator function represents the calculation of \emph{selecting one token from the sentence} and serves as a flexible extension of the binary indicator function $\mem{set}$. 
It returns a continuous similarity score between $-1$ and $1$ for a token, depending on its similarity with the tokens in the sentence.
Specifically, $\mem{vectors}(\vec{a}_i, \m{B})$ quantifies to what extent the $i$-th token vector $\vec{a}_i$ in sentence $A$ is semantically included in sentence $B$ by taking the maximum cosine similarity between $\vec{a}_i$ and the token vectors in $B$:
\begin{align}
    \mem{vectors}(\vec{a}_i, \m{B}) 
    = \max_{\vec{b}_j \in \m{B}} \cos(\vec{a}_i, \vec{b}_j)
    \in [-1, 1] \label{eq:member1}\text{,}
\end{align}
where $\cos(\vec{a}_i, \vec{b}_j)$ is the cosine similarity between $\vec{a}_i$ and $\vec{b}_j$.

\paragraph{Limitations of BERTScore's Indicator Function}
The indicator function $\mem{vector}$ lies at the heart of BERTScore, playing a crucial role in the computation of $P_{\mathrm{BERT}}$, $R_{\mathrm{BERT}}$, and $F_{\mathrm{BERT}}$. However, a critical limitation arises from BERTScore's reliance on maximum cosine similarity for its indicator function, which severely hinders its ability to capture the rich and nuanced meanings conveyed by a sentence.
\autoref{fig:concept} starkly illustrates this limitation through the visualization of BERTScore's $\mem{vectors}$ calculation. Consider the sentence \textit{We are the king and queen}, which evokes a broader, more abstract concept of \textit{royalty} through the co-occurrence of \textit{king} and \textit{queen}. When seeking an alignment for \textit{royalty}, BERTScore's indicator function heavily favors the token with the highest cosine similarity --- in this case, \textit{king}. This approach leads to a severe alignment bias towards the meaning of a single word, while the complex and implicit meanings conveyed by the entire sentence are overlooked.

\paragraph{SubspaceBERTScore}
To overcome the limitations of BERTScore regarding the expressiveness of its indicator function, we propose SubspaceBERTScore. 
This method extends BERTScore by employing the concept of subspace-based sentence representation and indicator function.

\paragraph{Extension of $P$, $R$, $F$ with Subspaces}
Based on the above discussions, we propose \emph{SubspaceBERTScore}, which calculates BERTScore's $R$, $P$, $F$ using the subspace representation of sentences and the subspace indicator function:
\begin{align}
    R_{\mathrm{subspace}} &= \frac{1}{|A|} \sum_{\vec{a}_i \in A}\softin(\vec{a}_i, \Subs{B}) \text{,} \\
    P_{\mathrm{subspace}} &= \frac{1}{|B|} \sum_{\vec{b}_i \in B}\softin(\vec{b}_i, \Subs{A}) \text{,} \\
    F_{\mathrm{subspace}} &= 2\frac{P_{\mathrm{subspace}} \cdot R_{\mathrm{subspace}}}{P_{\mathrm{subspace}} + R_{\mathrm{subspace}}} \text{,}
\end{align}
where $R_{\mathrm{subspace}}$, $P_{\mathrm{subspace}}$, and $F_{\mathrm{subspace}}$ are the final evaluation measures of SubspaceBERTScore.

\paragraph{Weighting by Importance}
Previous study~\cite{DBLP:conf/acl/BanerjeeL05,DBLP:conf/cvpr/VedantamZP15} has shown that infrequently occurring words play a more important role in sentence similarity than general words.
We apply importance weightings to our method as follows:
\begin{align}
    R_{\mathrm{subspace}} &= \frac{\sum_{\vec{a}_i \in A} \mathrm{weight(\vec{a_i})} \softin(\vec{a}_i, \Subs{B})}{\sum_{\vec{a}_i \in A} \mathrm{weight(\vec{a_i})}} \text{,} \\
    P_{\mathrm{subspace}} &= \frac{\sum_{\vec{b}_i \in B} \mathrm{weight(\vec{b_i})} \softin(\vec{b}_i, \Subs{A})}{\sum_{\vec{b}_i \in B} \mathrm{weight(\vec{b_i})}} \text{,}
\end{align}
where $\mathrm{weight}(\cdot)$ is a weighting function. 
We use the L2 norm of the vector \cite{DBLP:conf/emnlp/YokoiTASI20,DBLP:conf/emnlp/OyamaYS23}.

\begin{algorithm}[t]
\caption{Subspace indicator function $\softin(\v{w}, \Subs{A})$}
\label{alg:softind}
\begin{algorithmic}
    \Require $\ms{A} \in \mathbb{R}^{k \times d}$: Bases of $\Subs{A}$
    \Require $\v{w} \in \mathbb{R}^{1 \times d}$: A word vector
    \Ensure $\sigma \in \mathbb{R}$: Membership degree
    \If{$k=0$}
        \State \Return 0
    \Else
        \State $\widetilde{\v{}}_\w{w} \leftarrow \frac{\v{w}}{|| \v{w} ||} \in \mathbb{R}^{1 \times d}$
        \State $\m{U}^\top \in \mathbb{R}^{k \times k}, \sigma \in \mathbb{R}, \m{V} \in \mathbb{R} \leftarrow \textsc{SVD}(\ms{A} \widetilde{\v{}}_\w{w}^{\!\!\top})$\! 
        \State \Return $\sigma \in \mathbb{R}$
        \Comment{The output of $\softin(\v{w}, \Subs{A})$ is always \textbf{non-negative} because it is a singular value}
    \EndIf
\end{algorithmic}
\end{algorithm}

\section{Semantic Textual Similarity Task}
\label{sec:sts}
In this section, we examine the effectiveness of SubspaceBERTScore through the semantic textual similarity task~\cite[STS;][]{DBLP:conf/semeval/AgirreCDG12}.

\paragraph{Task}
An STS task calculates the similarity between two sentences.
For the STS evaluation protocol, we follow \citet{DBLP:conf/emnlp/GaoYC21}.
Its evaluation is based on the correlation between the similarity calculated by the model and corresponding human judgments.
We used datasets from the SemEval shared task 2012-2016 \cite{DBLP:conf/semeval/AgirreCDG12,DBLP:conf/starsem/AgirreCDGG13,DBLP:conf/semeval/AgirreBCCDGGMRW14,DBLP:conf/semeval/AgirreBCCDGGLMM15,DBLP:conf/semeval/AgirreBCDGMRW16}, STS benchmark \cite[STS-B;][]{DBLP:conf/semeval/CerDALS17}, and SICK-Relatedness \cite[SICK-R;][]{DBLP:conf/lrec/MarelliMBBBZ14}.
We used Spearman's $\rho$.

\paragraph{Embeddings}
We used 768-dimensional BERT$_\mathrm{base}$\footnote{\url{https://huggingface.co/bert-base-uncased}} \cite{DBLP:conf/naacl/DevlinCLT19}, which was pre-trained with BookCorpus and Wikipedia.
We used hidden states in the last layer.

\begin{table*}[t]
\centering
\scalebox{0.8}[0.8]{ 
\begin{tabular}{lccllllllllllll}
\toprule
\textbf{Method} & \textbf{Metric} & \textbf{Weighting} & \textbf{STS12} & \textbf{STS13} & \textbf{STS14} & \textbf{STS15} & \textbf{STS16} & \textbf{STS-B} & \textbf{SICK-R} & \textbf{Avg.} \\
\midrule
\midrule
CLS-cos & - & - & .215 & .321 & .213 & .379 & .442 & .203 & .424 & .314 \\ 
Avg-cos & - & - & .309 & .599$^\bigstar$ & .477 & .603 & .637 & .473 & .582$^\bigstar$ & .526 \\ 
WMD & - & - &  .238 & .443 & .389 & .531 & .532 & .384 & .509 & .432 \\
WRD & - & - &  .241 & .502 & .410 & .573 & .573 & .421 & .527 & .464 \\
DynaMax & - & - & .322 & .518 & .432 & .616 & .639 & .452 & .560 & .506 \\
\midrule
\midrule
\multirow{3}{*}{BERTScore} & $F$ & - & .312 & .546 & .450 & .602 & .636 & .446 & .553 & .506 \\
& $P$ & - & .261 & .532 & .462 & .576 & .622 & .443 & .559 & .494 \\
& $R$ & - & .350 & .527 & .416 & .602 & .623 & .430 & .522 & .496 \\
\midrule
\multirow{3}{*}{SubspaceBERTScore} & $F$ & - & \textbf{.335} & \textbf{.573} & \textbf{.476} & \textbf{.610} & \textbf{.650} & \textbf{.479} & \textbf{.562} & \textbf{.526} \\
& $P$ & - & \textbf{.282} & \textbf{.550} & \textbf{.488} & \textbf{.580} & \textbf{.630} & \textbf{.475} & \textbf{.568} & \textbf{.511} \\
& $R$ & - & \textbf{.369}$^\bigstar$ & \textbf{.552} & \textbf{.436} & \textbf{.611} & \textbf{.639} & \textbf{.462} & \textbf{.530} & \textbf{.514} \\
\midrule
\midrule
\multirow{3}{*}{BERTScore} & $F$ & L2 & .321 & .540 & .452 & .613 & .640 & .454 & .558 & .511 \\
& $P$ & L2 & .274 & .529 & .468 & .589 & .627 & .450 & .565 & .500 \\
& $R$ & L2 & .348 & .520 & .414 & .610 & .624 & .437 & .524 & .497 \\
\midrule
\multirow{3}{*}{SubspaceBERTScore} & $F$ & L2 & \textbf{.342} & \textbf{.568} & \textbf{.477} & \textbf{.621} & \textbf{.653}$^\bigstar$ & \textbf{.486}$^\bigstar$ & \textbf{.568} & \textbf{.531}$^\bigstar$ \\
& $P$ & L2 & \textbf{.292} & \textbf{.547} & \textbf{.492}$^\bigstar$ & \textbf{.592} & \textbf{.634} & \textbf{.479} & \textbf{.574} & \textbf{.516} \\
& $R$ & L2 & \textbf{.367} & \textbf{.544} & \textbf{.434} & \textbf{.620}$^\bigstar$ & \textbf{.640} & \textbf{.468} & \textbf{.532} & \textbf{.515} \\

\bottomrule
\end{tabular}
}
\caption{
A comprehensive comparison of similarity metrics in the STS task. The scores are Spearman's $\rho$.
The methods with the highest values, using the same pre-trained embeddings, are highlighted in $\bigstar$. Scores that showed improvement from BERTScore are denoted in \textbf{bold}.
}
\label{tab:result_sts1}
\end{table*}

\paragraph{Baselines}
We compared our method \textbf{SubspaceBERTScore} with other baseline similarity metrics.
The baselines included \textbf{Avg-cos} \cite{DBLP:conf/iclr/AroraLM17}, the cosine similarity between the averaged vectors, \textbf{CLS-cos} \cite{DBLP:conf/emnlp/GaoYC21}, the cosine similarity between the $[CLS]$ representations of the pre-trained language model, \textbf{DynaMax} \cite{DBLP:conf/iclr/ZhelezniakSSMFH19}, a set similarity based on fuzzy sets, Word Mover’s Distance \cite[\textbf{WMD};][]{DBLP:conf/icml/KusnerSKW15}, a metric based on optimal transport cost, and Word Rotator’s Distance \cite[\textbf{WMD};][]{DBLP:conf/emnlp/YokoiTASI20}, an optimal transport-based metric that improves WMD.

\paragraph{Main results}
The results are shown in \autoref{tab:result_sts1}.
In comparison to BERTScore, our method achieves superior correlation with human judgments across all three key metrics: F-score, precision, and recall. 
An important observation is that the performance consistently improves by subspace representation of the set.
The results suggest that simply replacing the representation of embedding sets and the indicator function with subspace-based alternatives significantly enhances our ability to capture and express the depth of linguistic semantics.

We also conduct an experiment using L2 norm as a weighting factor for the indicator function. 
This method has previously been proven effective in the STS task~\cite{DBLP:conf/emnlp/YokoiTASI20}. 
We see that both our proposed method and BERTScore improve their performance underlining the effectiveness of this weighting approach in both cases. 
Notably, our proposed method continued to outperform BERTScore even when L2 norm was used for weighting.

Our similarity also outperforms the fuzzy-set based similarity of DynaMax. 
This result suggests that the proposed subspace-based approach represents a set and set operations better than the fuzzy set-based approach in embedding space.

\section{Text Concept Set Retrieval Task}
\label{sec:tcsr}
In this section, we evaluate the capability of our proposed set operations ($\cap$, $\cup$, and $\in$) in effectively representing word sets. 

\paragraph{Task}
We evaluate our set operations by the set expansion task introduced by \citet{DBLP:conf/nips/ZaheerKRPSS17}.
In this task, the model is given a set of words that share a common concept or theme. 
The objective is to expand this set by retrieving relevant words from a vocabulary that fit the same concept. 
For instance, if the initial set includes words like ``apple'', ``banana'', and ``peach'', the task would be to identify and add other fruit names (e.g., ``orange'') to this set.
For the evaluation, we follow \citet{DBLP:conf/nips/ZaheerKRPSS17}.
We report recall (\textbf{R@k}) and \textbf{Median}, that indicate whether the words in the test set can be ranked higher. 

\paragraph{Embeddings}
We used the most standard pre-trained word embeddings in all of our experiments: 300-dimensional GloVe\footnote{\url{https://nlp.stanford.edu/projects/glove/}} \cite{DBLP:conf/emnlp/PenningtonSM14}, which was pre-trained with Common Crawl, and 300-dimensional word2vec\footnote{\url{https://code.google.com/archive/p/word2vec/}} \cite{DBLP:conf/nips/MikolovSCCD13}, which was pre-trained with Google News.

\paragraph{Set Expansion with Subspace Indicator Function}
To illustrate our subspace-based set expansion method (\textbf{Subspace Set}), we consider a set of fruit-related words. For example, let's take $\set{S}_\mathrm{fruit} = \{\w{apple}, \w{banana}, \dots\}$. This set is divided into two subsets: a 'span' subset used for creating a subspace representation, and a 'test' subset for evaluation. Let's assume $\w{orange} \notin \set{S}_\mathrm{fruit\_span}$ is a target word for testing.
(1) From the 'span' subset, we generate a subspace: $\Subs{Fruit} = \mathrm{span}(\set{S}_\mathrm{fruit\_span})$. For instance, if $\set{S}_\mathrm{fruit\_span} = \{\w{apple}, \w{banana}, \dots\}$, then $\Subs{Fruit} = \mathrm{span}(\v{apple}, \v{banana}, \dots)$.
(2) We define a subspace indicator function, which computes the degree to which a word vector belongs to the subspace. For a word $\w{w}$, the membership score is calculated as $\mathrm{score} = \softin(\v{orange}, \Subs{Fruit})$. This score reflects the extent to which $\w{w}$ aligns with the semantic characteristics of the subspace. 
This method effectively expand the set $\set{S}_\mathrm{fruit}$ by identifying words that share semantic properties with the subspace defined by the initial set.

\paragraph{Baselines}
We compared several baselines, which don’t require training on word sets, to our method.
\textbf{Random} just selects words randomly from the dataset’s vocabulary.
A simple unsupervised baseline with word embeddings uses the nearest neighbors in the embedding space (\textbf{Near})\footnote{
While \citet{DBLP:conf/nips/ZaheerKRPSS17} does not provide details about this method, we have inferred through our replication experiments that it uses a method based on the cosine similarity between the query word vector and other vectors to obtain the nearest neighbor.}.
We also compare a method based on fuzzy sets \cite[\textbf{Fuzzy set};][]{DBLP:conf/iclr/ZhelezniakSSMFH19} with our method.
Similar to our method, their method is designed to exploit both the flexibility of word vectors and rich set operations.
Fuzzy Set represents word set $\set{A}$ by max-pooled word vectors $\vec{s} = \max_{\w{w} \in \set{A}} \v{w}$. 
One major difference from our method is that Fuzzy set represents a set of word vectors by compressing them into a vector of fixed dimensions.
Although the Text Concept Set Retrieval task requires computing the degree of a word’s membership for a word set, their method does not provide it.
We instead used cosine similarity $\cos(\v{w}, \vec{s})$ between word vector $\v{w}$ of word $\w{w} \in \set{V}$ and $\vec{s}$ as the degree of membership to apply fuzzy sets to the task.

\paragraph{Dataset}
We used a previously created dataset \cite{DBLP:conf/nips/ZaheerKRPSS17}, which was denoted by ``LDA-1k, Vocab = 17k.'' in the paper.
The dataset ($\mathbf{D^{\mathrm{Set}}}$) contains 100 word sets, each of which consists of 50 words sampled from a common topic\footnote{This work used Latent Dirichlet Allocation \cite[LDA;][]{DBLP:journals/jmlr/BleiNJ03} as a topic model.}. 
Five pre-determined words from each set were used as the word set $\set{S}$.
An additional 800 word sets were used to train the models that require training on word sets.
\autoref{tab:dataset2} shows an example of the data and the number of test sets.

\begin{table}[tbp]
\setlength{\tabcolsep}{0.76mm} 
\centering
\scalebox{0.8}[0.8]{ 
\begin{tabular}{llllllllll}
\toprule
\textbf{Dataset ({\# Set})} & \multicolumn{5}{c}{\textbf{Example}}  \\
\cmidrule(l){2-6}
 & \textbf{Set} & \multicolumn{4}{c}{\textbf{Words (set elements)}} \\
\midrule
$\mathbf{D^{\mathrm{Set}}}$ (100) & $\set{S}_{3}$ & daily & news & paper & $\dots$ \\
\midrule
\multirow{3}{*}{$\mathbf{D^{\mathrm{Union}}}$ (100)} 
& $\set{S}_{12}$ & rider & bike & bicycle & $\dots$ \\
& $\set{S}_{51}$ & island & fishing & sea & $\dots$ \\
& $\set{S}_{12} \cup \set{S}_{51}$ & races & cycling & islands & $\dots$ \\
\midrule
\multirow{3}{*}{$\mathbf{D^{\mathrm{Intersect}}}$ (100)} 
& $\set{S}_{9}$ & tour & open & golf & $\dots$  \\
& $\set{S}_{72}$ & poker & casino & gambling & $\dots$ \\
& $\set{S}_{9} \cap \set{S}_{72}$ & money & won & player & $\dots$ \\
\bottomrule
\end{tabular}
}
\caption{
Examples from original dataset (denoted as $\mathbf{D^{\mathrm{Set}}}$) and additional $\mathbf{D^{\mathrm{Union}}}$ and $\mathbf{D^{\mathrm{Intersect}}}$ sets.
} 
\label{tab:dataset2}
\end{table}

\begin{table}[t]
\setlength{\tabcolsep}{1.3mm} 
\centering
\scalebox{0.8}[0.8]{ 
\begin{tabular}{lllcrrr}
\toprule
& \textbf{Method} & \textbf{Emb.} & \textbf{Set} & \textbf{R@100} & \textbf{R@1k} & \textbf{Med.} \\ 
\midrule 
\multirow{6}{*}{\rotatebox{90}{$\mathbf{D^{\mathrm{Set}}}$}} 
& Rand$^\spadesuit$ & - & $\x$ & 0.6 & 5.9 & 8520 \\
& Near$^\spadesuit$ & word2vec & $\x$ & 28.1 & 54.7 & 641 \\ 
& Fuzzy set & word2vec & $\chk$ & 19.9 & 47.2 & 1240 \\ 
& Fuzzy set & GloVe & $\chk$ & 30.9 & 69.0 & 320 \\ 
& Subspace set & word2vec & $\chk$ & 29.7 & 58.9 & 478 \\ 
& Subspace set & GloVe & $\chk$ & \textbf{35.7} & \textbf{72.7} & \textbf{246} \\
\midrule 

\multirow{6}{*}{\rotatebox{90}{$\mathbf{D^{\mathrm{Union}}}$}} 
& Rand & - & $\x$ &  0.6 & 6.0 & 8422 \\
& Near & word2vec & $\x$ & 17.5 & 34.3 & 3270 \\ 
& Fuzzy set & word2vec & $\chk$ & 2.8 & 17.1 & 4426 \\
& Fuzzy set & GloVe & $\chk$ & 5.4 & 32.0 & 2347 \\
& Subspace set & word2vec & $\chk$ & 18.4 & 46.9 & 1202 \\
& Subspace set & GloVe & $\chk$ & \textbf{24.4} & \textbf{68.3} & \textbf{407} \\
\midrule

\multirow{6}{*}{\rotatebox{90}{$\mathbf{D^{\mathrm{Intersect}}}$}} 
& Rand & - & $\x$ & 0.2 & 6.6 & 7929 \\
& Near & word2vec & $\x$ & 23.5 & 40.8 & 3304 \\
& Fuzzy set & word2vec & $\chk$ & 4.7 & 20.9 & 3420 \\
& Fuzzy set & GloVe & $\chk$ & 32.5 & 75.0 & 255 \\
& Subspace set & word2vec & $\chk$ & 25.7 & 45.7 & 1445 \\
& Subspace set & GloVe & $\chk$ & \textbf{44.2} & \textbf{83.7} & \textbf{149} \\
\bottomrule
\end{tabular}
}
\caption{Results of set retrieval task on $\mathbf{D^{\mathrm{Union}}}$ (top half) and $\mathbf{D^{\mathrm{Intersect}}}$ (bottom half). 
The ``Emb.'' column indicates which pre-trained embedding is used.
The ``Set'' column indicates whether each method is based on set computations: $\chk$ for incorporating set operations.
}
\label{tab:result_tcsr}
\end{table}

To evaluate the union and intersection sets, we prepared additional data through the union and intersection operations on two randomly-selected word sets from the original word sets ($\mathbf{D^{\mathrm{Set}}}$)\footnote{Note that these were based on the dataset LDA-1k, which was automatically generated by \citet{DBLP:conf/nips/ZaheerKRPSS17} using LDA, so the quality depends on their method.}.
The number of words in each set in $\mathbf{D^{\mathrm{Union}}}$ was limited to 50 to match the original dataset ($\mathbf{D^{\mathrm{Set}}}$). 
The number of words in each set in $\mathbf{D^{\mathrm{Intersect}}}$ was set to a minimum of 10.
Finally, 100 unions and intersections were randomly selected from these word sets with zero elements excluded.
See \autoref{tab:dataset2} for examples and statistics of the datasets.

\paragraph{Results}
In experiments on union and intersection, we compared our method only with Fuzzy Set.
The proposed method and Fuzzy Set can induce representations for the union and intersection using set operations defined in the word embedding space; the others cannot do so directly.
\autoref{tab:result_tcsr} shows the experimental results. 
Here our subspace-based set operation method (Subspace set) is the best among the methods that did not require training.
The results suggest that combining off-the-shelf pre-trained embeddings with appropriate set-oriented operations makes linguistic computation on sets feasible without additional training.
The results in $\mathbf{D^{\mathrm{Union}}}$ and $\mathbf{D^{\mathrm{Intersect}}}$ show that the our method outperform Fuzzy Set in most metrics.
As methods for achieving set operations in vector spaces, the proposed method is empirically more promising than the existing fuzzy set-based method.

\section{Related Work}
Symbol-based similarities between word sets have been proposed, such as Jaccard coefficient \cite{jaccard1901,DBLP:books/daglib/0001548,thada2013comparison} and TF-IDF-based similarity \cite{jurafsky2000speech}.
Unfortunately, symbol-based methods cannot capture the semantic similarity of similar sets or words when the symbols are different.

While many studies have explored representing word sets in pre-trained embedding spaces~\cite{DBLP:conf/icml/KusnerSKW15,DBLP:conf/emnlp/YokoiTASI20}, they primarily focus on set similarity. 
Our approach, however, extends beyond this by developing a comprehensive framework for various set operations within these spaces. 
Utilizing subspace properties, our method not only represents word sets but also performs a range of versatile operations, such as calculating textual similarities and membership degrees. 

Many methods for learning the representation of sets have been proposed because of the wide range of possible applications~\cite{DBLP:conf/nips/ZaheerKRPSS17,DBLP:conf/ijcai/PellegriniTFPJ21,DBLP:conf/icml/LeeLKKCT19,DBLP:journals/corr/VilnisM14,DBLP:conf/acl/AthiwaratkunW17}.
In contrast, our approach does not require additional training.
This enables us to compute set representations and operations using popular general-purpose language models, which are trained on the general domain \cite{DBLP:conf/nips/BrownMRSKDNSSAA20}.

\section{Conclusion}
This study introduces a novel framework for set representation and operations within pre-trained embedding spaces, employing linear subspaces grounded in quantum logic. 
This approach extends the scope of conventional embedding set operations by incorporating vector-based representations. 

\section*{Ethical Considerations}
We recognize the importance of addressing the inherent biases in pre-trained models, such as gender stereotypes. In our experiment, we used RoBERTa, which has gender biases~\cite{DBLP:journals/corr/abs-2105-05541}. 
We used this model in its original state to preserve the experimental conditions of BERTScore, acknowledging that such biases may influence our results.
However, we would like to emphasize that the focus of our work, which lies in sentence similarity, does not inherently add to or magnify these ethical concerns. 

\section*{Limitations}
Our SubspaceBERTScore is built upon the foundation of BERTScore, which presents a limitation in that our results and findings are inherently dependent on the characteristics and performance of BERTScore. While we chose BERTScore due to its robustness and popularity in the field, potential biases or shortcomings intrinsic to BERTScore might be incorporated into our extension. Nevertheless, this constraint also suggests future research possibilities, such as applying our subspace-based approach to other base sentence similarity metrics, further expanding the versatility and applicability of our method.

The experiments we conducted were exclusive to BERT and RoBERTa. Testing our methodology with other pre-trained models, like GPT-3~\cite{DBLP:conf/nips/BrownMRSKDNSSAA20}, could broaden its applicability and establish its robustness across various pre-trained models.

We evaluated our methodology primarily using English datasets. This decision was made to streamline our initial explorations rather than due to an inherent language-specific bias in our approach. We expect that our subspace-based methodology will be effective across various languages.

\section*{Acknowledgments}
This work was supported by JSPS KAKENHI Grant Number 22H03654, 22H03651, and JST, ACT-X Grant Number JPMJAX200S, Japan.

\bibliography{custom}

\newpage
\appendix
\section{Pseudocodes}
The pseudocodes for our SubspaceBERTScore and basic set operations are shown in Algorithms \ref{alg:union}, \ref{alg:intersection}, \ref{alg:complement}, and \ref{alg:subspacebertscore}.

\begin{algorithm}[h]
\caption{Union}
\label{alg:union}
\begin{algorithmic}
    \Require $\ms{A} \in \mathbb{R}^{k \times d}$: Bases of $\Subs{A}$
    \Require $\ms{B} \in \mathbb{R}^{\ell \times d}$: Bases of $\Subs{B}$
    \Ensure $\ms{A \cup B} \in \mathbb{R}^{r \times d}$: Bases of $\Subs{A} \cup \Subs{B}$
    \State $\m{M} \in \mathbb{R}^{(k + \ell)\times d} \leftarrow \textsc{concat\_rows}(\ms{A}, \ms{B})$
    \State $\ms{A \cup B} \in \mathbb{R}^{r \times d} \leftarrow \textsc{ortho\_normal}(\m{M})$
    \Comment{$r$ is rank of $\m{M}$}
    \State \Return $\ms{A \cup B}$
\end{algorithmic}
\end{algorithm}

\begin{algorithm}[h]
\caption{Intersection}
\label{alg:intersection}
\begin{algorithmic}
    \Require $\ms{A} \in \mathbb{R}^{k \times d}$: Bases of $\Subs{A}$
    \Require $\ms{B} \in \mathbb{R}^{\ell \times d}$: Bases of $\Subs{B}$ ($\ell \geq k$)
    \Require $\alpha$: Threshold below which the cosine of canonical angles is considered zero.
    \Ensure $\ms{A \cap B} \in \mathbb{R}^{m \times d}$: Bases of $\Subs{A} \cap \Subs{B}$
    \State $\m{M} \in \mathbb{R}^{k \times \ell} \leftarrow \ms{A} \ms{B}^{\top}$
    \State $\m{U} \in \mathbb{R}^{k \times \ell}, \m{\Sigma} \in \mathbb{R}^{\ell \times \ell}, \m{V}^\top \in \mathbb{R}^{\ell \times \ell} \leftarrow \textsc{SVD}(\m{M})$
    \\
    \Comment{$\m{\Sigma} = \mathrm{diag}(\sigma_1,\dots,\sigma_\ell)$ ($\sigma_1\geq\dots\geq\sigma_\ell$) has cosines of the canonical angles between $\Subs{A}$ and $\Subs{B}$.}
    \State $\m{W} \in \mathbb{R}^{m \times d} \leftarrow \m{U}[:, 1:\sigma_m]$
    \\
    \Comment{$m$ is the maximum index that satisfies $\abs{\sigma_i - 1} \leq \alpha$.} 
    \State \Return $\ms{A \cap B}$
\end{algorithmic}
\end{algorithm}

\begin{algorithm}[h]
\caption{Complement}
\label{alg:complement}
    \begin{algorithmic}
        \Require $\ms{A} \in \mathbb{R}^{k \times d}$: Bases of $\Subs{A}$
        \Ensure $\ms{\overline{A}} \in \mathbb{R}^{(d-k) \times d}$: Bases of $\Subs{\overline{A}}$
        \State $\m{U} \in \mathbb{R}^{d \times d}, \m{\Sigma} \in \mathbb{R}^{d \times k}, \m{V}^{\top} \in \mathbb{R}^{k \times k} \leftarrow \textsc{SVD}(\ms{A}^{\top})$
        \State $\ms{\overline{A}} \in \mathbb{R}^{(d-k) \times d} \leftarrow (\m{U}[:, k:d])^{\top}$
        \State \Return $\ms{\overline{A}}$
    \end{algorithmic}
\end{algorithm}

\begin{algorithm}[t]
\caption{SubspaceBERTScore}
\label{alg:subspacebertscore}
\begin{algorithmic}
\Require $\{\vec{a}^{(1)}, \ldots, \vec{a}^{(k)}\} \subseteq \mathbb{R}^{1 \times d}$: Token vectors of first sentence
\Require $\{\vec{b}^{(1)}, \ldots, \vec{b}^{(\ell)}\} \subseteq \mathbb{R}^{1 \times d}$: Token vectors of second sentence 
\Ensure $P \in \mathbb{R}, R \in \mathbb{R}, F \in \mathbb{R}$: Similarity score

\State $\m{A} \in \mathbb{R}^{k\times d} \leftarrow \textsc{stack\_rows}(\vec{a}^{(1)}, \ldots, \vec{a}^{(k)})$
\State $\m{B} \in \mathbb{R}^{\ell \times d} \leftarrow \textsc{stack\_rows}(\vec{b}^{(1)}, \ldots, \vec{b}^{(\ell)})$

\State $\m{A}_{\textrm{orth}} \in \mathbb{R}^{n \times d} \leftarrow (\textsc{ortho\_normal}(\m{A}^{\top}))^{\top}$
\State $\m{B}_{\textrm{orth}} \in \mathbb{R}^{m \times d} \leftarrow (\textsc{ortho\_normal}(\m{B}^{\top}))^{\top}$
\Comment{Orthonormalize the bases. $n$ and $m$ are the ranks.}

\State $\displaystyle R \leftarrow \sum_{i = 1}^k f(a_i) \softin(\vec{a}^{(i)}, \m{B}_{\textrm{orth}}) ~ \Big/ \sum_{i = 1}^k f(a_i)$ 
\State $\displaystyle P \leftarrow \sum_{j = 1}^\ell f(b_i) \softin(\vec{b}^{(j)}, \m{A}_{\textrm{orth}}) ~\Big/ \sum_{j = 1}^\ell  f(b_i)$ 
\Comment{$f(\cdot)$ is a weighting function.}

\State $F \leftarrow  2 (P \cdot R) ~/ (P + R)$
\State \Return $P, R, F$

\end{algorithmic}
\end{algorithm}

\section{WMT Results}

\begin{table*}[t]
\centering
\begin{tabular}{@{}lccccccccc@{}}
\toprule
\textbf{Method} & \textbf{Metric} & \textbf{cs-en} & \textbf{de-en} & \textbf{et-en} & \textbf{fi-en} & \textbf{ru-en} & \textbf{tr-en} & \textbf{zh-en} & \textbf{Avg.} \\
\midrule
\multirow{3}{*}{BERTScore} & F & .404 & .550 & .397 & .296 & .353 & .292 & .264 & .365 \\
                           & P & \textbf{.387} & .541 & .389 & .283 & .345 & .280 & .248 & .353 \\
                           & R & .388 & .546 & .391 & .304 & .343 & .290 & .255 & .360 \\
\midrule
\multirow{3}{*}{SubspaceBERTScore} & F & \textbf{.411} & \textbf{.557} & \textbf{.403} & \textbf{.309} & \textbf{.358} & \textbf{.303} & .264 & \textbf{.372} \\
                                    & P & .382 & \textbf{.548} & \textbf{.393} & \textbf{.290} & \textbf{.352} & \textbf{.294} & .248 & \textbf{.358} \\
                                    & R & \textbf{.391} & \textbf{.547} & \textbf{.392} & \textbf{.313} & \textbf{.358} & \textbf{.292} & \textbf{.259} & \textbf{.365} \\
\bottomrule
\end{tabular}
\caption{Comparison of BERTScore and SubspaceBERTScore on WMT18 for X-to-English translation tasks.}
\label{table:WMT18_results}
\end{table*}

An example of a practical task where our proposed method can be directly utilized is the automatic evaluation of machine translation systems. 
We applied our SubspaceBERTScore to the WMT18 ~\cite{DBLP:conf/wmt/MaBG18}.
The table \ref{table:WMT18_results} presents the results of applying both BERTScore and our proposed SubspaceBERTScore to various X-to-English translation settings in the WMT18 competition.
Our proposed method consistently outperforms BERTScore. 
Specifically, for all X-to-English translation settings, we observed an increase in Kendall’s $\tau$ for F-score, Precision (P), and Recall (R).
It is evident that SubspaceBERTScore shows an improvement across all the considered metrics, underscoring the effectiveness of our proposed method in evaluating machine translation systems.

\section{Vector vs. Set of Vectors}
\autoref{tab:analysis} presents the comparison results on the STS-B dataset. We utilized embeddings from BERT and SimCSE~\cite{DBLP:conf/emnlp/GaoYC21}. 
Avg-cos and CLS-cos are based on the cosine similarity of vectors, employing methods that compress information into a single vector. In contrast, BERTScore among other methods is based on sets of vectors.
A key takeaway from these results is the suggestion that methods using sets of vectors outperform those using a single vector in evaluating similarity. 
It is particularly interesting that even SimCSE, which aims to optimize average cosine similarity, shows superior score when using set-based similarity.

\begin{table}[t]
\centering
\begin{tabular}{llc}
\toprule
\textbf{Emb.} & \textbf{Method} & \textbf{STS-B} \\
\midrule
\multirow{4}{*}{BERT-base} & CLS-cos           & .203\\
                           & Avg-cos           & .473 \\ 
                           & BERTScore$^\dag$         & .446 \\ 
                           & SubspaceBERTScore$^\dag$ & .479 \\ 
\midrule
\multirow{4}{*}{SimCSE}    & CLS-cos           & .769 \\
                           & Avg-cos           & .786 \\ 
                           & BERTScore$^\dag$         & .798 \\ 
                           & SubspaceBERTScore$^\dag$ & .801\\ 
\bottomrule
\end{tabular}
\caption{Comparison of vector-based similarity and set-based similarity metrics ($\dag$) on the STS-B dataset.}
\label{tab:analysis}
\end{table}

\end{document}